\documentclass{article}

\usepackage{PRIMEarxiv}

\usepackage[utf8]{inputenc} 
\usepackage[T1]{fontenc}    
\usepackage{hyperref}       
\usepackage{url}            
\usepackage{booktabs}       
\usepackage{amsfonts}       
\usepackage{nicefrac}       
\usepackage{microtype}      
\usepackage{lipsum}
\usepackage{fancyhdr}       
\usepackage{graphicx}       
\graphicspath{{media/}}     
\usepackage{multirow}
\usepackage{enumitem}
\usepackage{tabularx}
\usepackage{longtable}
\title{A Comprehensive Review of Visual-Textual Sentiment Analysis from Social Media Networks
}

\author{
  ISRAA KHALAF SALMAN AL-TAMEEMI, Mohammad-Reza Feizi-Derakhshi \\
  Computerized Intelligence Systems Laboratory  \\
  Department of Computer Engineering\\
  University of Tabriz \\
  Tabriz, Iran\\
  \texttt{\{asra\_salman, mfeizi\}@tabrizu.ac.ir} \\
   \And
  Saeed Pashazadeh, Mohammad Asadpour \\
  Department of Computer Engineering    \\
  University of Tabriz \\
  Tabriz, Iran  \\
  \texttt{\{pashazadeh, m\_asadpour\}@tabrizu.ac.ir} \\
  \And
  Corresponding author: Mohammad-Reza Feizi-Derakhshi (e-mail: mfeizi@tabrizu.ac.ir).
}

\begin{document}
\maketitle

\begin{abstract}
Social media networks have become a significant aspect of people’s lives, serving as a platform for their ideas, opinions and emotions. Consequently, automated sentiment analysis (SA) is critical for recognising people’s feelings in ways that other information sources cannot. The analysis of these feelings revealed various applications, including brand evaluations, YouTube film reviews and healthcare applications. As social media continues to develop, people publish vast quantities of information in various formats, like text, pictures, audio, and video. Thus, traditional SA algorithms have become limited, as they do not consider the expressiveness of other modalities. By including such characteristics from various material sources, these multimodal data streams provide new opportunities for optimising the expected results beyond text-based SA. Our study focuses on the forefront field of multimodal SA, which examines visual and textual data posted on social media networks. Many people are more likely to utilise this information to express themselves on these platforms. To serve as a resource for academics in this rapidly growing field, we introduce a comprehensive overview of textual and visual SA, including data pre-processing, feature extraction techniques, sentiment benchmark datasets, and the efficacy of multiple classification methodologies suited to each field. We also provide a brief introduction of the most frequently utilised data fusion strategies and a summary of existing research on visual–textual SA. Finally, we highlight the most significant challenges and investigate several important sentiment applications.
\end{abstract}


\section{Introduction} \label{sec:introduction}
Today's affordable and comprehensive Web provides plenty of valuable social data to enhance decision-making. Sentiment analysis (SA) has become increasingly important for tracking people’s emotions and opinions by evaluating unorganised, high-dimensional, multimodal and noisy social data. SA aids in gathering information about a given subject or topic in different fields, such as business, marketing, and intelligence services. It is an area of natural language processing (NLP) that concentrates on analysing massive amounts of online material generated by users and posted regularly on social networking sites such as Flickr, Instagram, Facebook and Twitter. These networks contain a wide range of social content, enabling the detection of sentiment reflected by textual and visual data. Nowadays, if one wishes to purchase a consumer product, one does not have to rely solely on the guidance of friends and family because many user reviews and comments about the product are available in public Web forums. For example, we may consult Amazon reviews to determine which product best suits our needs. The significance of reviews arises from their ability to correctly represent previous consumer experiences. However, identifying and monitoring online opinion sites and collecting the information contained inside seems to be a difficult task. A typical human reader has difficulty finding relevant websites, extracting their thoughts and summarising them; hence, automated SA tools are necessary. In addition to real-world applications, several application-oriented research articles on topics such as stock market forecasting \cite{Li2014}, political election prediction \cite{Kagan2015} and even healthcare \cite{Yadav} have been published. In Li et al. \cite{Li2014}, financial news articles were used to measure their influences on stock price return. A sentiment analysis strategy for forecasting the result of presidential elections in a Twitter nation was reported by Kagan et al. \cite{Kagan2015}. Social media data was utilised by Yadav et al. \cite{Yadav} to build a patient-assisted system via medical SA. As a result, significant work has been conducted on user SA using textual data for numerous applications, which is commonly referred to as the process of automatically evaluating whether a part of the text is positive, negative, or neutral.
    
In addition to textual data, images play an important role in expressing users’ emotions on various online social media platforms. With the emergence of social networks, images have become a simple and effective method of conveying information to online users. Similar to textual data, images transmit various levels of emotion to their viewers. Nevertheless, extracting and interpreting an image’s sentiment remains a challenge. As a result, visual sentiment analysis (VSA) has become important for investigating online multimedia content. It involves the ability to recognise an object, action, scene and emotional content from an image and classify them into different sentiment polarities, which can be performed by portraying the image in terms of colour and utilising gist features to categorise images based on different emotions. Moreover, deep learning (DL) architectures have seen tremendous success in the computer vision field \cite{Panagakis2021}. Convolutional neural networks (CNNs) have demonstrated robust and accurate feature learning capabilities compared with DL-based networks. CNN models are more similar to human performance in visual recognition; they have been used successfully in tasks such as feature learning \cite{Yu2021}, sentence classification \cite{Zhang2018b} and image classification \cite{Xin2019}. Given the rapid growth and popularity of social media, scholars have been able to broaden the scope of SA to include other interesting applications. For example, the vast majority of VSA studies focus on close-up photographs of faces, in which facial expressions are analyzed in order to deduce their feelings and predict their reactions \cite{Zhang2019}.
    
As previously mentioned, most of the prior research on textual SA has been widely investigated [61–122], revealing various methodologies for mining opinion, including machine learning (ML), lexicon-based and hybrid techniques using Twitter text data. Moreover, only a few relevant studies are available in the literature on VSA \cite{ShaikAfzal2021,Jia2012,Amencherla2017,Wu2020a,Yang2014,Li2018,Jou2014,Desai2020,Chen2020,Song2018,Wu2020,Cetinic2019,Yang2018,Ou2021,Yadav2020}. However, most of the published research has focused on analysing data from a single modality, neglecting the complementary information included in visual and textual content. In addition, human expressions are quite complicated because words, visuals and their combinations can be used to represent a vast spectrum of feelings, most of which need context to be comprehended properly. Thus, understanding the inner relationship between image and text is important because this combination can change or improve the meanings or sentiments.
    
Multimodal SA has recently received considerable attention from academics to cope with diverse social media trends, given that people are increasingly expressing themselves and sharing their experiences through multiple types of social media. The basic goal of multimodal SA is to capture users’ feelings by combining expressive data from many modalities. It also focuses on building semantic linkages between user-provided contents, such as text, images, videos and other types of media. The current state of research in multimodal SA is divided into three primary categories:
\begin{itemize}
    \item Social media image and tagged content analysis \cite{Zhao2019}.
    \item Analysis of audio and visual information \cite{Li2020b}.
    \item Human–human and human–machine interaction analysis \cite{Rozanska2019}.
\end{itemize}
Our review article will focus on the most significant field of multimodal SA, which works on visual and textual information posted on social media platforms. Many people are more inclined to use these data to express themselves on these platforms. The visual and textual characteristics will be integrated to establish the general attitudes conveyed in the postings, which will then be categorised as positive, negative or neutral.
\subsection{SCOPE OF THIS SURVEY} \label{subsec:scope_survey}
Most previous surveys have ignored the majority of visual–textual SA techniques, focusing instead on supervised ML and DL techniques. Although the present work discusses these approaches, it differs from earlier research by providing a more comprehensive examination of SA because it covers many aspects of the discipline, such as challenges, applications, sentimental datasets, feature extraction techniques, data fusion approaches and classification strategies. Scholars and novices will benefit greatly from this study because they may obtain extensive knowledge about this topic from just one paper. Our study varies from previous studies by detailing the benefits and drawbacks of the most commonly used fusion approaches, which may help researchers select the best answer to their problems. The following are some of the survey’s most important contributions:
\begin{itemize}
    \item It reviews current works to offer researchers a complete understanding of the methodologies and resources available for visual and textual SA.
    \item It provides a comprehensive overview of visual and textual SA, including data pre-processing, feature extraction techniques and sentiment benchmark datasets.
    \item This study categorises and summarises the most common SA methodologies, namely, ML, lexicon-based, hybrid and DL methods.
    \item It provides a brief introduction of the most widely used data fusion strategies and summarises the existing research on visual–textual SA by referencing previously published works.
    \item It summarises the applications and challenges associated with SA.
\end{itemize}
The remainder of this paper is organised as follows: In Section II, sentiment definitions and terms are introduced. Section III presents the basic architecture of the visual–textual SA and thoroughly discusses the complete process, including critical steps, such as data pre-processing, feature extraction techniques and the most important SA classification algorithms for textual and visual components, along with the most important measures used in this area. Section IV presents SA challenges, and Section V discusses the significant sentiment applications. Section VI concludes this paper.
\subsection{MOTIVATION} \label{subsec:motivation}
There has been a significant increase in the quantity of publications in SA and opinion mining, making it one of the most explored topics. SA can assess, identify and classify people’s emotions, attitudes, feelings and opinions conveyed in the text. As a direct result of the rapid expansion of social networking sites (e.g. Twitter, Instagram and Facebook), multimedia information like text, images, audio and videos is crucial in conveying people’s thoughts. Multimodal SA derived from social media is becoming a topic of discussion in many fields of artificial intelligence, such as computer vision and pattern recognition. However, most current SA research focuses on a single modality, making it difficult to discern people’s sentiments. The accuracy, reliability and robustness of these single-mode systems are limited. Thus, the usage of several modalities should be investigated to improve SA systems.

This study aims to investigate practical and empirical aspects of blog data accessible through social media networks (e.g. Twitter, Instagram, Facebook and other websites) that provide such multimedia data. The topic of SA has become a topic of interest due to the growing expansion of information on social media. Several facets of multimedia business studies are investigated, all of which benefit the business society as a whole. Another aim is to develop a mathematical framework for multimodal SA, which includes numerous inputs to improve SA’s overall performance. Once implemented, the model will assist social media researchers in determining numerous sentiment-related factors. Our study focuses on providing a comprehensive description of the datasets, feature extraction methods, fusion approaches, classification methods and difficulties encountered while performing multimodal SA.
\section{SENTIMENT DEFINITIONS AND TERMS} \label{sec:sentiment_definition}
In NLP, the terms ‘affect’, ‘feeling’, ‘emotion’, ‘sentiment’, and ‘opinion’ are often interchanged. ‘Emotion’ has several descriptions and related concepts, many of which rely on Scherer’s emotion theory \cite{Scherer2005}. Scherer contends that emotions are transient phenomena that demand activation. They are composed of cognitive appraisals, physical responses, action inclinations (e.g. fight or flight), facial and vocal expressions and subjective experiences. The mood is a long-lasting term, broad emotional condition that occurs for no apparent reason and remains for hours or days. Feelings are emotional experiences that are unique to each individual.

Deonna and Teroni \cite{Deonna2012} defined affective phenomena philosophically; they defined sentiment as a deeply held belief that may take many forms, such as ‘the passion you may have for your hamster, your allegiance to your nation, your contempt for the financial system and your extreme liking for the latest technological item’. Sentiment manifests only when the sentiment holder is met with an entity or object. According to  Deonna and Teroni, emotional interactions with an object or entity may be utilised to identify or monitor sentiment. For example, if A Likes B, then A believes that B is in a good mood.

Munezero et al. \cite{Munezero2014} comprehensively discussed emotion recognition and SA in text to distinguish between emotion and sentiment based on duration. Emotions fade quickly, whereas sentiments stay for a long time. The authors described opinions as debatable and ambiguous assessments that do not have to be emotionally charged. SA is often associated with opinion mining, and it evaluates highly emotional attitudes and feelings \cite{Bhadane2015}. In sum, different from emotions, sentiments, and opinions may not always appear in behaviours or expressions. Emotions include a person, subjective feelings, physical changes and a target, whereas sentiments have a sentiment holder, polarity (positive or negative) and an object. Previously, there existed an interest in understanding others’ ideas and impressions. However, more organisations have shown interest in comprehending consumer attitudes to better satisfy their needs and enhance client acquisition and retention \cite{Mantyla2018}.
\section{VISUAL-TEXTUAL SA FRAMEWORK} \label{sec:visual_textual}
\begin{figure}
\centering
\includegraphics[width=\textwidth]{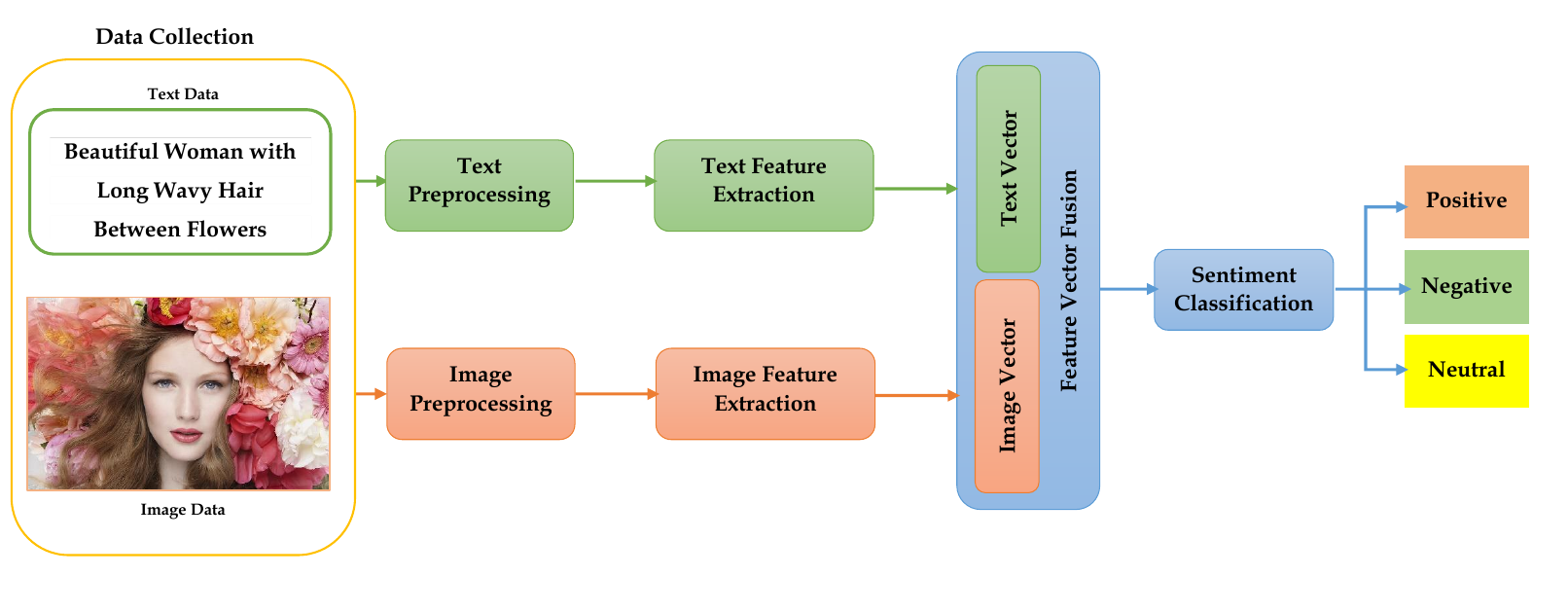}
\caption{Basic architecture for visual–textual sentiment analysis.}\label{fig1}
\end{figure}

The basic architecture for visual–textual SA is shown in Figure \ref{fig1}. It comprises different modules, starting with the data collection module, which is the first phase in visual–textual SA. The data are in the form of texts, images from a dataset or social media posts. During the pre-processing phase, the data obtained from the data collection module is prepared for additional processing. Features are taken from each modality separately since each one has distinct characteristics and must be handled separately and represented as vectors throughout the feature extraction and vector representation phases. These feature vectors are then combined to generate a single feature representation for recognising the emotion of the user’s content. The combined feature vector is used in the classification phase to classify the data collected and processed into positive, negative or neutral. The last module of this design visualises and shows user attitudes based on the previous study. We will provide a brief review of each of these phases in the following sections.

\subsection{TEXTUAL SENTIMENT ANALYSIS} \label{subsec:textual_sentiment_analysis}
SA, through social media content, aims to capture the sentiment information from social media posts. SA for social networks is more difficult than extracting sentiment from webpages (e.g. blogs and forums) due to the unique features of social media data. The brief phrasing and unstructured style of social media communication pose various issues. This text analysis allows businesses to determine client needs and political parties to conduct opinion surveys. Many applications rely on textual sentiment research, including stock market forecasting, box office and election predictions. Three conventional textual SA levels are used: document, sentence and entity levels. Social media comments are usually no more than two lines long due to space constraints. In this scenario, document and sentence levels are the same. Therefore, two levels of textual SA need to be defined for social media, including message/sentence and entity levels \cite{Giachanou2016}. This section outlines the various pre-processing operations, feature extraction techniques and computational approaches related to textual SA.
\subsubsection{TEXT PRE-PROCESSING} \label{subsubsec:text_preprocessing}
Information obtained from different resources, particularly social media, is frequently unstructured. Raw data may be noisy and include spelling and grammatical errors \cite{Liu2012}. Thus, texts must be cleaned up before analysing them. Given that many words are meaningless and add nothing to the text (e.g. stop words, prepositions, punctuation and special characters), pre-processing seeks to improve the analysis and minimise the dimensionality of the input data. Several typical tasks are included during the entire procedure as follows.
\begin{itemize}
    \item Lowercasing. Changing all of the texts’ letters to the same case. The word 'Ball', for instance, will become 'ball'.
    \item Substituting for negative words. Tweets include various negational concepts. The negation process converts 'won’t' and 'can’t' to 'will not' and 'cannot', respectively.
    \item Eliminating unnecessary information [including punctuation, hashtags (\#), special characters (\$, \&, \%, …), additional spaces, stop words (e.g. ‘the’, ‘is’, ‘at’, etc.), URL references, @username, numbers and non-ASCII-English letters are omitted to maintain the uniqueness of the information encoded in English]. Such information does not expect the expression of users’ emotions.
    \item Translating emoticons. Nowadays, users utilise emoticons to express their thoughts, feelings and emotions. As a result, translating all emoticons to their corresponding words will produce better results.
    \item Changing words with repeated characters to their English origins. Individuals often employ words with repeated letters (e.g. 'coooool') to convey their feelings. 
    \item Expanding acronyms to their original words using an acronym dictionary. Abbreviations and slang are poorly constructed words that are frequently used in tweets. They must be restored to their original form.
    \item Tokenisation. This stage deconstructs text into tokens, which are small textual units (e.g. documents into phrases, sentences into words).
    \item Part-of-speech (PoS) tagging. Many structural components in a text, like verbs, nouns, adjectives, and adverbs, are identified during this stage.
    \item Lemmatisation. It is the process of reducing a particular word to its simplest form, identical to stemming but it retains word-related information like PoS tags.
\end{itemize}
Several researchers have taken advantage of the pre-processing techniques. Zainuddin et al. \cite{Zainuddin2018} used text pre-processing approaches, including stemming and stop word removal, to analyse attitudes in texts. Sharma et al. \cite{Sharma2020a} used PoS tagging to perform text-based SA on an online movie review dataset. Normalisation, URL removal and acronym expansion have also been considered. Sailunaz et al. \cite{Sailunaz2019} used these approaches to demonstrate how text preparation improves Twitter SA accuracy. Pradha et al. \cite{Pradha2019} performed text-based SA on a Twitter dataset using lemmatisation and stemming.
\subsubsection{TEXT FEATURE EXTRACTION} \label{subsubsec:text_feature_extractiong}
Feature extraction, also known as feature engineering, represents a critical part of the SA process since it directly influences the effectiveness of sentiment classification \cite{Kasri2019,Avinash2019}. This assignment aims to gather essential data (e.g. phrases representing emotions) from the text that explains important features of the text. The three types of feature extraction methods are as follows.

A set of hand-crafted features was introduced by Mohammad et al. \cite{Mohammad2013}. These features comprise the following:
\begin{itemize}
    \item All Caps. The number of words that have all letters capitalised.
    \item Emoticons. A simple method of describing characteristics that considers whether emoticons are present (positively or negatively) and if the emoticon serves as the final segmentation unit.
    \item Elongated Units. The percentage of total words that are extended (words that repeat a single character three times or more; for example, ‘gooood’).
    \item Sentiment Lexicon. Various sentiment lexicons may be used to capture textual features by calculating the number of sentiment words, the score of the final sentiment word, the overall sentiment score and the maximum sentiment score for each lexicon.
    \item Cluster. The overall number of terms contained within each of Twitter's 1,000 different clusters using the NLP tool \cite{Gimpel2011}.
    \item Ngrams (terms of presence and frequency). It is the most basic feature encoding commonly used in information retrieval and SA. Features are defined as a single word or a list of $n$ consecutive words that may be combined to make a unigram, bigram or trigram, and the frequency counts of those words. The term ‘presence’ assigns a binary value to the words (zero if the word is present, and one otherwise).
    \item PoS Tags. It represents the labels that define the function of words in a language. Words may be divided into numerous components of speech  (e.g. verbs, nouns, adjectives, articles, adverbs, pronouns, prepositions, conjunctions and interjections). For instance, ‘The phone is beautiful’ will be attached with a Stanford PoS tagger presented by Toutanova et al. \cite{Toutanova2003}: The (determiner DT), phone (noun NN), is (verb VBZ), beautiful (adjective JJ). Adjectives are used as important opinion indicators in some SA methods \cite{A.2016}.
    \item Opinion Words and Phrases. People often use opinion words to convey their positive or negative feelings (e.g. nice and amazing represent positive feelings; horrible and dreadful represent negative feelings). 
    \item Negations. Negative words can modify and reverse the emotional polarity (also known as valence or opinion shifters \cite{Aggarwal2012,Loia2014}). The most frequent negative words are ‘not’, ‘never’, ‘none’, ‘nobody’, ‘nowhere’, ‘neither’ and ‘cannot’. However, during the pre-processing stage, these terms are often regarded as stop words and are excluded from the text. Negation words should be treated with caution due to their importance because not all negated words result in negation.
\end{itemize}

Statistical Feature Representation. Here, input text is converted into a feature vector of fixed length that can be used in classification procedures. This method analyses text word weights using predefined keywords and generates a digital vector representing the text’s feature vector. Some significant text representation approaches are as follows.
\begin{itemize}
    \item Bag-of-Words (BoW) model. It is one of the most basic and widely used strategies for translating text to numerical form (vector) \cite{Kasri2019}. Nevertheless, it loses the textual syntactic information because it ignores word order, sentence structure or grammatical construction, focusing instead on word occurrence. The BoW model builds a vocabulary of all unique terms in the document, where the vocabulary length equals the number of distinct words. Then, it encodes each phrase as a fixed-length vector where the value of each point in the vector reflects the count or frequency of each word in the training set. The "term frequency-inverse document frequency (TF–IDF)" is a straightforward and effective extension of the BoW.
    \item TF–IDF. It is a statistical measure of the importance of a word to a document within a corpus of texts. This is achieved by multiplying two metrics: the number of times a term appears in a document (TF) and its inverse document frequency (IDF) over a group of documents. The TF–IDF value grows according to the number of times a word occurs in the document and is offset by the number of corpus documents that include the term, which helps to account for the fact that particular words occur more frequently than others. The TF–IDF is often computed using the scikit-learn library’s vectoriser class.
\end{itemize}

Distributed Word Representation. In contrast to the BoW model, the distributed word representation of a meaningful concept (such as a word, paragraph, or document) spreads the information regarding the concept along a vector. As a result, each point in the vector might indicate a non-zero value for a particular idea. Typically, distributed word representations (word embeddings) are used with DL models. The most commonly used distributed word representations are presented as follows.
\begin{itemize}
    \item Word2vec. It is a popular method for obtaining distributed word representations; it was created by Mikolov et al. \cite{Mikolov2013} utilising naive neural networks (NNs). Its architecture comprises the continuous BOW (CBOW) model, which estimates the current word based on the surrounding context words, and the Skip-Gram (SG) model, which uses the current word to estimate the context words.
    \item Global Vectors (GloVe). It is an unsupervised learning approach proposed by Pennington et al. \cite{Pennington2014} to create word embeddings from the corpus by collecting global word–to-word co-occurrence matrices to reveal important linear substructures of the word vector space. Due to its parallel implementation, the GloVe model may be trained efficiently on bigger datasets \cite{Yadav2020a}.
    \item FastText. It is a novel technique based on Bojanowski’s SG model \cite{Bojanowski2017} that treats each word as a bag of character n-grams. Each letter n-gram has a vector representation, with words encoded as the sum of these representations. It is rapid, allowing the models to be quickly trained on big corpora and word representations to be calculated for terms not in the training data.
\end{itemize}

Several researchers have used different types of feature extraction methods to retrieve the most important textual information. Jabreel and Moreno \cite{Jabreel2019} proposed a DL model that solved the emotion classification problem in Twitter data by exploiting the word2vec method to extract the tweet features. To extract text features, Kaibi et al. \cite{Kaibi2019} utilised feature extraction models, namely, word2vec, a prominent word embedding model based on NN architectures; GloVe; and FastText models, to classify sentiment on a Twitter dataset using support vector machine (SVM). Ahuja et al. \cite{Ahuja2019} investigated the effects of TF–IDF and n-gram on SA. The author Mohey El-Din \cite{Mohey2016} proposed an upgraded BOW to analyse textual reviews from the CiteULike website. Mee et al. \cite{Mee2021} used regression and SA, specifically TF–IDF, to investigate the relationship between textual properties and Twitter user characteristics.
\subsubsection{TEXT SENTIMENT CLASSIFICATION APPROACHES} \label{subsubsec:text_sentiment_classification_approaches}
\begin{figure}
\centering
\includegraphics[width=\textwidth]{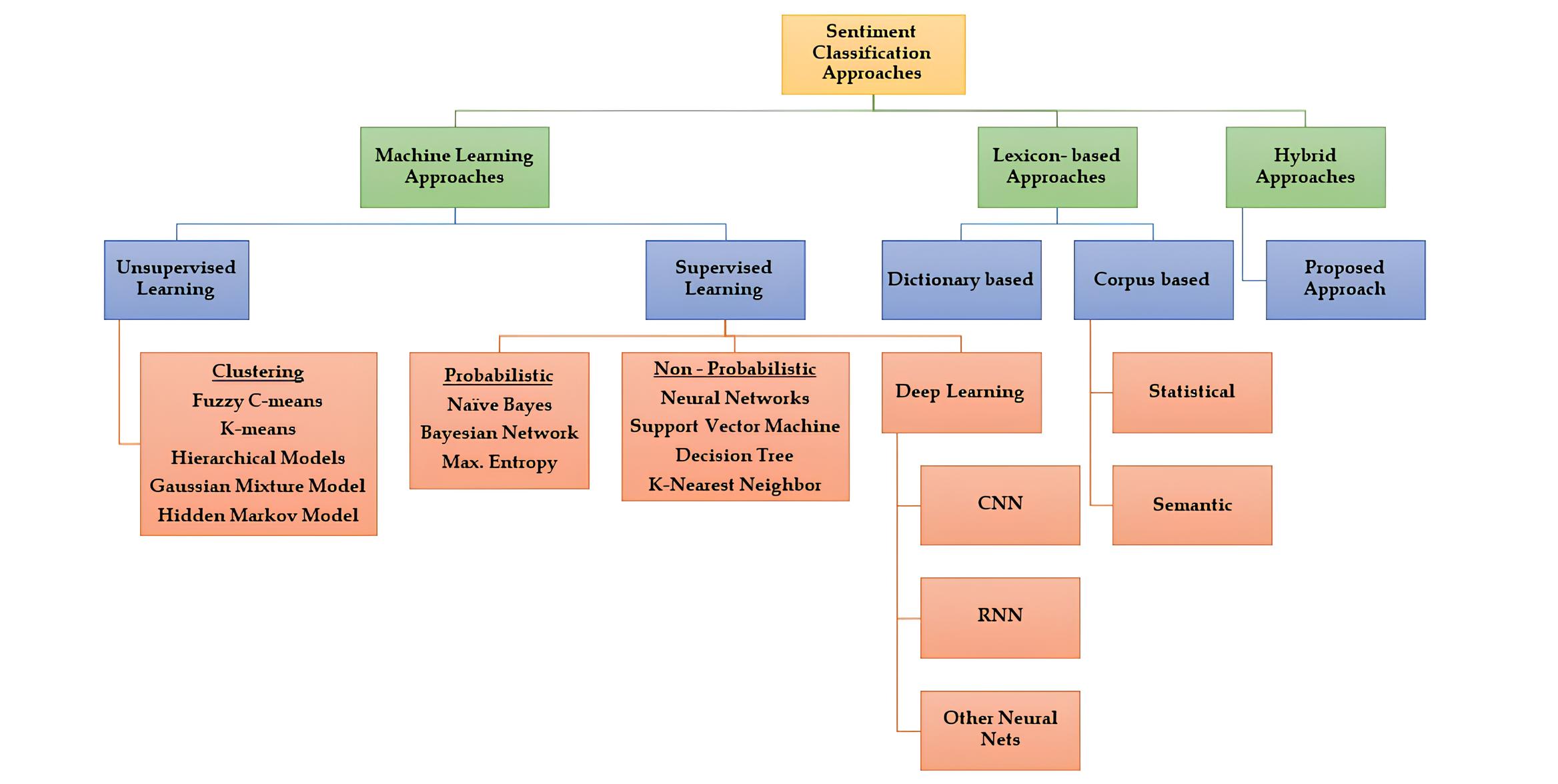}
\caption{Text sentiment classification approaches.}\label{fig2}
\end{figure}

SA is a promising research field with applications in various fields. As a result, researchers are always proposing, evaluating and comparing new methods to improve SA performance. Existing SA methods may be classified into ML, lexicon-based, hybrid and DL techniques. The most extensively used method is ML, which uses ML methods and linguistic properties to classify sentiments. The Lexicon-based approach utilises a sentiment lexicon that has already been predefined to get the sentiment value for a document by integrating the scores of all the terms in the document. The predefined lexicon contains frequently used words or phrases with their corresponding sentiment score. Hybrid methods integrate machine learning with lexicon-based techniques to address the weaknesses of each methodology and improve sentiment classification performance. DL algorithms use word embedding to extract the most relevant properties from the text, outperforming ML-based techniques for textual sentiment classification. This section briefly summarises the textual SA approaches shown in Figure \ref{fig2}. The most notable advantages and disadvantages associated with these approaches are listed in Table \ref{tbl1:adv-disadv_sentiments_analysis}.

\begin{table}
    \caption{Advantages and disadvantages of various sentiment analysis approaches}
    \begin{tabular}{p{70pt}p{190pt}p{190pt}}
    
    \hline
    \textbf{Method}  &\textbf{Advantages}  &\textbf{Disadvantages}   \\
    \hline
    ML  &\begin{itemize}[leftmargin=*]
        \item Dictionary is unnecessary
        \item Excellent classification accuracy
        \item High precision and adaptability
        \end{itemize}  &\begin{itemize}[leftmargin=*]
                        \item More time is required
                        \item Controlled by the domain
                        \item Requires human interaction and information labelling
                        \item Does not consider the sentiment information encoded in sentiment terms    
                        \end{itemize}   \\
    &   &   \\
    \hline
    Lexicon-based   &\begin{itemize}[leftmargin=*]
                        \item No labelling is required
                        \item Depends on the domain
                        \item Takes less time
                    \end{itemize}   &\begin{itemize}[leftmargin=*]
                                        \item Requires strong language resources
                                        \item Low precision
                                        \item Requires dictionaries with various viewpoints
                                        \item Ignore the sentence context
                                    \end{itemize}   \\
    &   &   \\
    \hline
    Hybrid Method   &\begin{itemize}[leftmargin=*]
                        \item Takes less time
                        \item Combines the advantages of several procedures
                        \item Can identify and quantify sentiment at the concept level
                        \item More accurate
                    \end{itemize}   &\begin{itemize}[leftmargin=*]
                                        \item Low precision
                                        \item Lacks reliability
                                    \end{itemize}   \\
    &   &   \\
    \hline
    DL  &\begin{itemize}[leftmargin=*]
            \item Features are automatically identified and optimised
            \item The same NN may handle several tasks and data types
            \item DL architecture is more adaptable to future issues
            \item It captures the nonlinear interaction between the user and the object, allowing for faster sentiment prediction
        \end{itemize}   &\begin{itemize}[leftmargin=*]
                            \item Requires an extensive number of labelled samples to get good results.
                            \item Expensive training due to complex data models
                            \item Long training time
                        \end{itemize}   \\
    \hline
    \end{tabular}
    \label{tbl1:adv-disadv_sentiments_analysis}
\end{table}
\subsubsection*{3.1.3.1) MACHINE LEARNING APPROACHES}
Machine learning approaches are one of the most extensively utilised techniques for categorizing textual information into different sentiment labels (e.g. negative, positive and neutral) using training and testing parts of the textual dataset. These techniques begin with the creation of a training set and labelling it with the sentiment. After a statistical analysis is performed on the training data, a collection of features is retrieved and sent to a classifier model. Once the classifier has been trained with sentiment labels, it will be able to estimate the sentiment polarity of an unannotated sample \cite{Devika2016}. According to \cite{Yusof2015}, these strategies can be categorized as  supervised \cite{Oneto2016} and unsupervised \cite{Li2017} learning techniques. The supervised strategy is used when the classification problem involves a specified set of classes. However, the unsupervised approach may prove the optimal solution when it is impossible to establish this set due to a lack of labelled data. Although ML approaches may recognize domain-specific structures in text, which improves classification performance, they usually require massive training datasets to achieve good performance. Moreover, the classifiers developed for one dataset may underperform in another \cite{Agarwal2016,Yoo2018}.
\subsubsection*{3.1.3.1.1) SUPERVISED LEARNING}
Supervised learning is established by utilising a labelled dataset to train the system with the labels denoting the classes (e.g. positive, neutral and negative). The classes are given several comparable sets of features and distinct labels. To categorise the training data, supervised learning approaches use probabilistic or non-probabilistic classification models. The probabilistic classifier makes a prediction based on the probability distribution across classes and calculates the likelihood that the extracted features belong to one of the classes (0 or 1). It often uses Bayes' theorem as its foundation \cite{Fisch2014}.The classifier performs classification using mixture models, with each class being an element of the mixture. Probabilistic classifiers are easy to build, computationally efficient compared with other techniques and require little training data. However, if the data do not completely match the distribution assumptions, then classification performance may suffer. In certain cases, the probabilistic classifier may be inadequate. The disadvantages of using a probabilistic classifier may be avoided by utilising a non-probabilistic technique. A non-probabilistic classification method does not display class propagation and is useful when probabilistic classifiers are ineffective. It distinguishes the feature space from the sample origin and returns the class associated with that space. The subsections that follow give an overview and comparison of the most commonly used supervised classification algorithms for SA.

\textbf{Naïve Bayes (NB)}. It is one of the most frequently used probabilistic text classifiers. The model is based on Bayes’ theorem for forecasting the probability of a given set of characteristics, and the input features are assumed to be naturally self-regulating. The overall probability is determined by multiplying the prior probability by the likelihood. This method simplifies the training and classification processes. Li et al. \cite{Li2020a} developed a danmaku sentiment dictionary and suggested an innovative method for analysing the feelings of danmaku reviews using a sentiment dictionary and NB. The approach is useful for monitoring the broad emotional orientation of danmaku videos and forecasting their popularity. Shrivastava et al. \cite{Shrivastava2021} developed a Twitter SA programme to analyse Republic of Indonesia’s presidential candidates in 2019. This SA method involves data collection using Python libraries, text processing, data training and testing. Finally, text classification is performed using an NB approach. Kunal et al. \cite{Kunal2018} suggested utilising Tweepy in conjunction with TextBlob, a Python framework, to process and categorize tweets through NB. Kharisma et al. \cite{Kharisma2021} used term weighting to obtain the best model combination. The tagged data were cleaned and transformed into structured data to be analysed. The pre-processed data were then categorised using multinomial and Bernoulli NB. The experiments showed that Bernoulli NB is more accurate than multinomial NB.

\textbf{Maximum Entropy (ME)}. It is a probabilistic classifier mostly used for text classification based on distribution uniformity. It depends upon the assumption of known data and ignores unknown data. This technique uses search-based optimisation to estimate the weights of the characteristics to predict the lowest risk while maximising entropy. Habernal et al. \cite{Habernal2014} conducted a performance evaluation of multiple classifiers to assess sentiment polarity, revealing that the ME classifier outperforms all other classifiers in predicting sentiment polarity on a given dataset. Htet and Myint \cite{Htet2018} created a social media (Twitter) data analysis system to determine the state of people’s health, education and business based on Twitter data, in which SA is utilised to propose these specified requirements using the ME classifier. On the basis of ME and NB, Ficamos et al. \cite{Ficamos2017} developed a sentiment classification approach that uses PoS tags to extract unigram and bigram features. They obtained greater estimation accuracy by concentrating on a certain issue. Hagen et al. \cite{Hagen2015} advocated combining SVM with other models, such as ME and stochastic gradient descent optimisation, compared with previous SemEval editions with different feature sets, the combined model was selected as the best performing model of the year.

\textbf{Bayesian Network (BN)}. It is a probabilistic graphical classifier that illustrates how random variables are related. The distribution of this model may be described simply as a directed acyclic graph, with vertices and edges representing variables and dependence relations, respectively. The network’s structure is easily expandable, allowing new variables to be added. BN can assist in the decision-making process when faced with a complex problem by assessing the probabilities of its causes and outcomes. Gutierrez et al. \cite{Gutierrez2019} studied the literature on the use and significance of BN in the SA area to assess textual emotions as part of research on text representation and BN. Wan et al. \cite{Wan2016} suggested an ensemble sentiment classification technique based on the majority vote principle of several classification methods, including NB, SVM, BN, C4.5 decision tree and random forest algorithms, to classify tweets about airline services.

\textbf{Support Vector Machine (SVM)}. It is a linear classifier that can manage discrete and continuous variables and divide data linearly or nonlinearly. It has a strong theoretical foundation and outperforms most other algorithms in terms of classification accuracy. The primary objective of an SVM classifier is to identify the optimal hyperplane for class separation. Hyperplanes with a large margin to a training point from either class are more efficient because they minimize the generalisation error of a classifier.\newline
SVM has been used successfully in a number of investigations. Al-Smadi et al. \cite{Al-Smadi2018} presented state-of-the-art methodologies that rely on supervised ML for dealing with the difficulties of Arabic hotel ratings. Deep recurrent NN (RNN) and SVM are developed and trained by utilising lexical, word, syntactic, morphological and semantic information. A review dataset of Arabic hotels has been used to evaluate the proposed methodologies; the SVM strategy performs better than the other deep RNN techniques. Nafis et al. \cite{MohdNafis2021} suggested an improved hybrid feature selection strategy to improve ML sentiment classification. Firstly, two customer review datasets were obtained and pre-processed. Next, they introduced a hybrid feature selection framework that uses TF–IDF and SVM–recursive feature elimination to assess and rank the features iteratively. Both approaches were developed and tested using the two datasets, and only the k-top features from the ranking features were utilised for sentiment classification. Finally, the proposed approach was evaluated using an SVM classifier. Obiedat et al. \cite{Obiedat2022} proposed a hybrid strategy integrating SVM, particle swarm optimisation and other oversampling algorithms to deal with unbalanced data. SVM has been utilised as an ML classification approach by improving the dataset of reviews from several restaurants in Jordan. Rezwanul Huq et al. \cite{Rezwanul2017} proposed two strategies to properly categorise the sentiment label from tweets, one based on k-nearest neighbour and the other on SVM. Naz et al. \cite{Naz2019} developed a technique that deals with Twitter sentiment classification using n-gram and SVM classifiers. They investigated the influence of weighting on classifier accuracy using three alternative weighting methods. They also used a sentiment score vector of tweets to enhance the performance of the SVM classifier.

\textbf{K-Nearest Neighbour (KNN)}. It is a non-probabilistic classifier that is easy to implement and works by comparing the new data points with the training points. It computes the Euclidean distance between the new point and its neighbours and assigns it to one of the classes based on the distance. Here, K represents the number of neighbours, and the new data point is then assigned to the class containing most of its neighbours. However, the algorithms become difficult if the number of independent variables exceeds a certain threshold. Naw Naw \cite{Naw2018} conducted sentiment classification on education, business, crime and health using KNN and SVM classifiers. The system provided educational authorities, economists, government organisations and health analysts with results related to these domains. Irfan et al. \cite{Irfan2018} conducted an emotional analysis of the 2013 curriculum. The KNN approach was used for sentiment classification. An ensemble of numerous feature sets was used, including textual, Twitter-specific, lexicon-based, PoS and BOW features. The experiment results revealed that ensemble features outperform individual features when it comes to sentiment classification. Damarta et al. \cite{Damarta2021} employed text mining to regulate the service quality in PT PLN (Persero) by categorising Twitter data using KNN. The obtained data were pre-processed and categorised as negative, neutral or positive. The results achieved 87.41\% accuracy using KNN. Muktafin et al. \cite{Muktafin2021} utilised KNN and TF–IDF algorithms with NLP to categorise conversations into ‘satisfied’ and ‘dissatisfied’ categories. Their results achieved 74\% accuracy.

\textbf{Decision Tree (DT)}. This technique uses a set of IF–Then rules to construct a tree-like structure with nodes and branches that are linked together, making it suitable for regression and classification problems. This network has a root node, decision nodes and leaf nodes. Decisions are made at decision nodes depending on the characteristics extracted from the dataset. DTs work admirably even when the input data volume is large. Phu et al. \cite{Phu2017} categorised positive, negative and neutral semantics for English texts using an ID3 approach of a DT. The semantic classification was based on several criteria produced by the ID3 algorithm on 115,000 English texts. The test was created using 25,000 English texts, and 63.6\% accuracy was obtained. Es-Sabery et al. \cite{Es-Sabery2021} developed a unique MapReduce-enhanced weighted ID3 DT classification algorithm for Opinion Mining, which includes three aspects. Firstly, n-gram or character-level, BOW, word embedding (GloVe and word2vec), FastText and TF–IDF feature extractors were used to effectively discover and collect important features from the provided tweets. Secondly, feature selections (e.g. chi-square, gain ratio, information gain and Gini index) were investigated to reduce the dimensionality of the high features. Thirdly, the characteristics were classified using an enhanced ID3 DT classifier that uses weighted information gain rather than the standard ID3 information gain. Fitri et al. \cite{Fitri2019} conducted SA in three stages: data pre-processing, classification, and evaluation, in order to determine the sentiment polarities (positive, negative, or neutral) in people's reviews on an anti-LGBT campaign in Indonesia. They developed different ML algorithms, including NB, DTs and RF, to perform the classification. The experiments showed good results related to each classifier.

\textbf{Neural Network (NN)}. It has lately gained popularity as a useful categorisation approach. It works by extracting features from linear datasets and modelling the outcome as a non-linear function of these characteristics. Three layers—input, output, and hidden—each comprising several ordered neurons, make up a typical NN design. The synaptic connections that take place between the neurons of each layer serve to establish a link between two layers that follow one another. In a gradient descent training procedure, each link is assigned a weight value, which is computed based on minimising an overall error function. There may be several hidden layers with a single input and output layer, where the number of output classes equals the number of output nodes. The feedforward NN is one of the most frequently used NNs, in which signals travel in one direction, from the input to the hidden layers and finally to the output layer. Bhargava et al. \cite{Bhargava2019} suggested a method for analysing tweets in one of the Indian languages (Hindi, Bengali or Tamil). To avoid overfitting and error accumulation, 39 sequential models were constructed with optimal parameter values using RNN, long short-term memory (LSTM) and CNN. Moraes et al. \cite{Moraes2013} proposed document-level sentiment classification to automate the classification of textual reviews on a particular subject as positive or negative. A standard BOW technique was used to extract all the important features, and the classification was performed using SVM, NB, and artificial neural network (ANN). In the experiments, the ANN outperformed the SVM on the benchmark dataset of movie reviews. Vinodhini and Chandrasekaran et al. \cite{Vinodhini2016} compared NN-based sentiment classification approaches (i.e. backpropagation NN, probabilistic NN (PNN) and homogeneous ensemble of PNN) employing varying degrees of word refinement as features for feature-level sentiment classification. The outcomes of ANN-based approaches were compared with two statistical methods using a dataset of Amazon product reviews. In classifying the sentiment of product reviews, the PNN outperformed the other two NN algorithms. The use of NN-based sentiment classification combined with principal component analysis as a feature reduction methodology shortens the training time. Table \ref{tbl2:adv-disadv_supervised_learning} details the advantages and disadvantages of probabilistic (i.e. NB, BN and ME) and non-probabilistic (i.e. SVM, KNN, DT and NN) classifiers.

\begin{table}
    \caption{Advantages and disadvantages of supervised learning classifiers.}
    \label{tbl2:adv-disadv_supervised_learning}
    \begin{tabular}{p{70pt}p{190pt}p{190pt}}
    \hline
    \textbf{Method}  &\textbf{Advantages}  &\textbf{Disadvantages}   \\
    \hline
    NB  &\begin{itemize}[leftmargin=*]
        \item Simple to implement and understand
        \item Low-resource computing demands
        \item Less data and training time than other approaches
        \end{itemize}  &\begin{itemize}[leftmargin=*]
                        \item Assumes characteristics are independent, which is seldom the case
                        \item Restricted by data scarcity because a probability value must be determined for each possible value.    
                        \end{itemize}   \\
    \hline
    BN   &\begin{itemize}[leftmargin=*]
                        \item Model construction takes less time because it is comprehensible even in complex areas.
                        \item Manages missing data and gets accurate results with limited training data
                        \item Prevents overfitting
                    \end{itemize}   &\begin{itemize}[leftmargin=*]
                                        \item Computationally costly; thus, it is rarely used
                                        \item Unsuitable for problems that contain many features
                                    \end{itemize}   \\
    \hline
    ME   &\begin{itemize}[leftmargin=*]
                        \item Useful when the previous distributions are unknown
                        \item Quick at obtaining information from text and dealing with massive amounts of data
                    \end{itemize}   &\begin{itemize}[leftmargin=*]
                                        \item Has a tendency towards overfitting
                                        \item Lacks reliability
                                    \end{itemize}   \\
    \hline
    SVM  &\begin{itemize}[leftmargin=*]
            \item Stable in high-dimensional spaces
            \item More accurate and simpler to train
            \item Memory-efficient due to high-dimensional kernel mapping
            \item Strong and can manage large, sparse collections of samples
        \end{itemize}   &\begin{itemize}[leftmargin=*]
                            \item Insufficient performance when the number of features exceeds the sample number
                            \item Requires a proper kernel function
                            \item Poor interpretability due to lack of probabilistic explanation
                        \end{itemize}   \\
    \hline
    KNN &\begin{itemize}[leftmargin=*]
            \item Simple to comprehend and apply
            \item Trains quickly
            \item Handles noisy data well
            \item Works well with samples that contain several class labels
        \end{itemize}   &\begin{itemize}[leftmargin=*]
                            \item When comparing unlabelled samples to a large number of possible neighbours, lazy learners suffer high computing costs
                            \item Sensitive to data local structure
                            \item Limited memory
                            \item Operates slowly due to its supervision
                        \end{itemize}  \\
    \hline
    DT  &\begin{itemize}[leftmargin=*]
            \item Easy and quick
            \item Yields accurate results
            \item Understandable representation
            \item Facilitates gradual learning
            \item Saves memory
            \item Handles noisy data
            \item Finds the optimal split attribute using entropy, Gini index and information gain
        \end{itemize}   &\begin{itemize}[leftmargin=*]
                            \item Requires extensive training
                            \item Due to replication issues, DTs may have considerably more complicated representations for some ideas
                            \item Overfitting issues
                        \end{itemize}   \\
    \hline
    ANN &\begin{itemize}[leftmargin=*]
            \item Handles complicated relationships between variables and greater generalisation even with noisy data
            \item Efficient for situations with high dimensionality
            \item Fast execution
        \end{itemize}   &\begin{itemize}[leftmargin=*]
                            \item Theoretically complicated and difficult to execute
                            \item High memory consumption
                            \item Requires more training time and a huge dataset in certain situations
                        \end{itemize}   \\
    \hline
    \end{tabular}
\end{table}
\subsubsection*{3.1.3.1.2) UNSUPERVISED LEARNING}
The fundamental limitation of supervised SA is the need for large annotated datasets to train a classification algorithm. Domain adaptation of supervised classifiers is difficult because certain domains need a more formal or lengthier text input (e.g. film reviews) that helps build the Treebanks dataset. The same approach cannot be used for social media data due to brief reviews (tweets are 140 characters in length). When the labelled data is unavailable, unsupervised SA is used. This approach does not require labelled data to produce sentiment predictions, which reduces the labelling costs. Unsupervised clustering finds a structure in the incoming data point and groups it with similar objects/points. Data points are assigned membership values via the fuzzy C-means method based on their distance from the cluster centre. A higher membership grade signifies closer proximity to the cluster’s core. In K-means clustering, the clusters have K randomly chosen centres. Each input data point is allocated to the nearest centre, and new centres are calculated. The algorithms finish when K does not change.

Many studies have used unsupervised approaches. Harish et al. \cite{Harish2014} proposed a new method based on term frequency vectors using clustering to encode text corpora. The mean and standard deviation of each cluster’s phrase frequency vectors were expressed symbolically (interval-valued). The term frequency vectors were clustered using a fuzzy C-means clustering algorithm based on an adaptable squared Euclidean distance among interval feature vectors. They used popular datasets (e.g. 20 Mini Newsgroup, 20 Large Newsgroup and Vehicles Wikipedia) as well as innovative datasets (e.g. Research paper abstracts and newsgroups on Google) to evaluate the suggested model’s performance. The results revealed that acquired classification accuracy outperforms existing methods. Orkphol and Yang \cite{Orkphol2019} used K-means to group similar microblog tweets indicating relevant sentiments about a product. Given the nature of microblogging, the dataset was sparse and high-dimensional. To solve this challenge, the authors used TF–IDF to choose important characteristics and singular value decomposition to decrease the high-dimensional dataset while keeping the most relevant features. The artificial bee colony (ABC) was used to determine the ideal starting state of centroids for K-means, the silhouette analysis method was used to determine the perfect K, and SentiWordNet was used to score each group after categorising them into K groups. Their technique showed its effectiveness in improving the K-means results. Han et al. \cite{Han2020} suggested a semi-supervised approach with numerous classifiers and a dynamic threshold to manage the shortage of initial annotated data. The training data were repeatedly auto-labelled using a dynamic threshold function. Moreover, the proposed weighted voting approach compared the performance of various SVM classifiers. The proposed model achieved maximum SA accuracy across datasets of various initial labelled training data sizes. Fernández-Gavilanes et al. \cite{Fernandez-Gavilanes2016} constructed an unsupervised dependency parsing-based text classification system that uses sentiment lexicons and NLP approaches. The Cornell Movie Review, Obama–McCain Debate and SemEval-2015 datasets showed the ’resilience and efficiency of the system. García-Pablos et al. \cite{Garcia-Pablos2018} described a W2VLDA system that can classify sentiment polarity, separate aspect terms from opinion words, and classify aspect categories for any language or domain. They used the SemEval 2016 task 5 (ABSA) dataset to assess the domain aspect and sentiment classification. Many industries (e.g. hotels, restaurants and electronics) and languages (e.g. English, Spanish, French and Dutch) had competing outcomes.

\subsubsection*{3.1.3.1.3) LEXICON-BASED APPROACHES}
In SA, lexicon-based approaches are one of the most basic methods for the study of sentiment. It uses an opinion lexicon (a predetermined set of words) that utilises scores to categorise words as negative or positive. A score may be a basic polarity quantity for positive, negative, or neutral terms (such as +1, -1, or 0) or a numerical value expressing emotional power or intensity. The semantic orientation values assigned to a document’s words determine its final orientation. When documents are tokenized into small words or short phrases, emotional scores are assigned to each element from the lexicon. The final sentiment can be calculated by summing or averaging each word's emotional scores. The lexicon-based technique works well for phrases and features. It is an unsupervised learning model since it does not need data for training. The fundamental issue with this strategy is domain dependency since words can have multiple interpretations and uses, rendering a good phrase bad in another. This problem can be resolved by developing a domain-specific sentiment lexicon or changing an already-existing vocabulary. The three major methods of collecting opinion words are manual-, dictionary- and corpus-based approaches. Manual procedures are time-consuming and are not always used. It is typically used as a last check after the other two automated processes to verify that no errors are made. According to the dictionary-based approach, synonyms have comparable emotional polarities, whereas antonyms have opposing emotional polarities. This method generates emotion lexicons utilising WordNet \cite{Miller1990} and the thesaurus \cite{Mohammad2009}. It starts by manually gathering known-orientation seed words. Synonyms and antonyms are then searched for each word in the list. The list is iterated until no new words are found. Some errors necessitate human intervention. Unlike dictionary-based approaches, corpus-based methods begin with a collection of initial sentiment words with a predefined orientation and then use syntax or co-occurrence characteristics to search across a huge corpus for additional emotional words that fit their orientation. It includes a statistical method for determining a word’s polarity based on how often it occurs in negative text and a semantic method for determining sentiment based on word similarity. Table \ref{tbl3:adv-disadv_sentiments_analysis} lists the advantages and disadvantages of the dictionary- and corpus-based approaches.

Many researchers have used lexicon-based approaches. Huang et al. \cite{Huang2020a} developed sentiment CNN to analyse sentences’ emotions using contextual and sentiment information from sentiment terms. Word embeddings give contextual data, whereas lexicons provide sentiment information. A highway network is created to allow for the adaptive integration of sentiment and contextual information from sentences by increasing the association between their characteristics and sentiment words. Yu et al. \cite{Yu2018} used an alternative sentiment embedding learning method to enhance the current pre-trained word vectors by building a model for word embedding refinement that uses real-valued sentiment intensity ratings from sentiment lexicons. Jurek et al. \cite{Jurek2015} introduced a new lexicon-based SA method for real-time tweet content analysis. The approach comprises two core parts: sentiment normalisation and an evidence-based mixture function, which are used to forecast sentiment intensity instead of positive/negative labelling and promote the usage of the combination of the sentiment classification procedure. Sanagar and Gupta \cite{Sanagar2020} developed a robust, unsupervised sentiment lexicon learning technique for new genres. The method and progressive learning are used to extract polarity seed concepts from a corpus of randomly selected source areas. Thereafter, the genre-level information is delivered to the target areas. Unlabelled training data from the source and target areas are utilised to construct a sentiment lexicon using latent semantic analysis. In a dictionary-based approach, Park and Kim \cite{Park2016} developed a dictionary-based strategy to create a thesaurus lexicon designed for sentiment classification. The proposed approach gathered a thesaurus utilising three online dictionaries and only stored co-occurrence keywords to enrich the reliability of the thesaurus lexicon. Also, this approach expanded the thesaurus vocabulary, which is a collection of synonyms and antonyms, to improve post accessibility and sentiment classification performance. Sanagar and Gupta \cite{Sanagar2016} provided a review paper discussing the polarity lexicon based on two parts: a review of the literature from ancient to current methodologies and a discussion of free-source polarity lexicons. Wang et al. \cite{Wang2015} developed an improved random subspace method called PoS-RS, a sentiment classification method based on parts of speech. This method regulates the balance between accuracy and variety of base classifiers by including two essential factors, namely, content and function lexicon subspace rates. Jha et al. \cite{Jha2018} constructed an emotion-aware dictionary using various domain data, including annotated data from the source domain and unlabelled data from the source and target domains. It is then used to identify unlabelled reviews in the target domain. The recommended strategy identifies 23\%–24\% more terms in the target domain than existing approaches. In the corpus-based approach, Agarwal and Mittal \cite{Agarwal2016a} proposed a method for SA using corpus-based semantic orientation. They investigated unique sentiment-bearing feature extraction strategies and conventional techniques, such as unigram, PoS-based and dependency features. In addition, they provided an approach for creating SA methods for areas with sparsely labelled data. Luo et al. \cite{Luo2013} developed a novel recommendation technique that incorporates the sentiment aspect of tagging into a social tag recommendation system. The tag sentiment data, which are provided in the form of users’ subjective polarity towards annotated resources, function as an additional information filter to offer users more desired and positive resources. Thus, sentiment tagging can improve recommendation efficiency.

\begin{table}
    \caption{Advantages and disadvantages of dictionary- and corpus-based approaches}
    \label{tbl3:adv-disadv_sentiments_analysis}
    \begin{tabular}{p{70pt}p{190pt}p{190pt}}
    \hline
    \textbf{Method}  &\textbf{Advantages}  &\textbf{Disadvantages}   \\
    \hline
    Dictionary-based Approach  &\begin{itemize}[leftmargin=*]
                                    \item Computationally affordable if dataset training is not required
                                    \item Represents a solid technique to quickly develop a vocabulary with many sentiment words and their orientation
                                \end{itemize}  &\begin{itemize}[leftmargin=*]
                                    \item Fails to discover sentiment terms with a certain domain orientation, making it inappropriate for context- and domain-specific categorization
                                    \item The compilation of dependency rules is a time-consuming
                                \end{itemize}   \\
    \hline
    Corpus-based Approach   &\begin{itemize}[leftmargin=*]
                                \item Simplicity
                            \end{itemize}   &\begin{itemize}[leftmargin=*]
                                                \item Requires a large dataset to identify word polarity and hence the textual sentiment
                                                \item Depends heavily on the polarity of the words in the training corpus because polarity is calculated for the terms in the corpus
                                            \end{itemize}   \\
    \hline
    \end{tabular}
\end{table}

\subsubsection*{3.1.3.1.4) HYBRID APPROACHES}
The hybrid technique combines lexicon- and ML-based approaches. It is principally inspired by the need to combine ML’s high precision and flexibility, with the robustness of a lexicon-based approach to deal with ambiguities and incorporate the context of sentiment words. Thus, the hybrid strategy may improve sentiment classification by combining both methodologies to overcome their disadvantages and maximise their advantages \cite{Yousif2019}. The true value of this technique is the presence of the lexicon/learning symbiosis, which can be utilised to enhance the efficacy and usability of the lexicon. It achieves high model accuracy by using strong supervised learning methods. The drawback is that the ideas expressed include many irrelevant terms related to the subject. Due to this noise, feelings are often assigned a neutral polarity rather than being detected as positive or negative \cite{DAndrea2015}. Ghiassi et al. \cite{Ghiassi2013} created a unique hybrid method by integrating n-grams and dynamic ANN. They constructed two classifiers, an SVM and an ANN, using emoticons and tweets as features. Khan et al. \cite{Khan2014} suggested a hybrid method for Twitter feed categorization. The suggested technique involves pre-processing the text before providing it to the classifier. Devi et al. \cite{Devi2019} classified documents using ML models that combined dictionaries and the HARN algorithm presented in a lexicon-based approach. The reviews were domain-classified using NB and SVM classification algorithms, and the polarity was then calculated at the document level using the HARN technique. The result showed that the hybrid method is better than the HARN algorithm by 80\%–85\%. Shin et al. \cite{Shin2016} developed a CNN that combines an attention mechanism and lexicon embeddings. They used word scores from various lexicons to create lexicon embeddings. The CNN model was built using separate convolution, multichannel concatenation, and nave concatenation. They showed that lexicon combination improves the performance, reliability and effectiveness of CNN. Elshakankery et al. \cite{Elshakankery2019} suggested a hybrid incremental learning method for Arabic tweet SA (HILATSA) uses ML and a lexicon-based approach for determining sentiment polarity in tweets. They classified and created a lexicon of words, emoticons, idioms, and other important lexicons using SVM, logistic regression and RNN. They studied the Levenshtein distance in SA to handle different word forms and spelling issues. HILATSA was tested using six datasets, and the results found that combining ML and lexicon-based methodologies in textual SA for social networks is effective.
\subsubsection*{3.1.3.1.5) DEEP LEARNING APPROACHES}
The implementation of ANN-based DL has considerably enhanced the SA field. DL is a rapidly growing topic in ML that provides techniques for comprehending the representation of the features in either a supervised or unsupervised form \cite{Rojas-Barahona2016}. DL is a concept that refers to NNs with numerous layers of perceptron motivated by the human brain \cite{Vateekul2016}. This architecture allows for training more complicated models on a considerably larger dataset, resulting in an outstanding performance across a broad range of application areas, including NLP, computer vision, and speech recognition \cite{Zhang2018a}.

DL comprises various NN models, including CNN, RNN and others that have been adopted in many research studies. Alayba and Palade \cite{Alayba2021} improved sentiment classification by combining CNNs and LSTM networks and removing CNN’s max-pooling layer, which reduces the length of the generated feature vectors by convolving the filters on the input data. As a result, the LSTM networks will receive vectors that have been sufficiently captured by the feature maps. The authors also evaluated many successful strategies for developing and expressing text features to increase the accuracy of Arabic sentiment classification. Salur and Aydin \cite{Salur2020} proposed a unique hybrid DL model that strategically integrates several word embeddings (i.e. word2vec, FastText and character-level embeddings) with various DL algorithms (i.e. LSTM, gated recurrent unit (GRU), bidirectional LSTM (BiLSTM) and CNN). The proposed model combines features from several DL word embedding techniques and classifies texts according to their emotion; it achieves better results in terms of sentiment classification. Zulqarnain et al. \cite{Zulqarnain2020} developed a unique two-state GRU and encoder approach called E-TGRU to construct an effective framework for SA. The findings showed that given sufficient training data, The GRU model is capable of effectively acquiring the words used in user opinions. The Results presented that E-TGRU outperformed GRU, LSTM and bi-LSTM using the Internet Movie Database (IMDB) and Amazon Product Review datasets. Li et al. \cite{Li2021} proposed an SA model for online restaurant reviews by integrating word2vec, bidirectional GRU and the attention technique. The findings revealed that the proposed model achieved excellent performance compared to other popular SA models in terms of overall performance. Sharma et al. \cite{Sharma2020} suggested word2vec word embedding and CNN for classifying short sentences. The method uses a pre-trained word2vec model to create word vectors and a CNN layer to extract improved features for sentence classification. Basiri et al. \cite{Basiri2020} proposed two deep fusion models based on a three-way decision theory to investigate drug reviews. When the deep method achieved low confidence in classifying the test samples, the first fusion model (3W1DT) was developed using a DL method as the main classifier and a traditional learning method as the secondary classifier. In the second fusion model (3W3DT), three extremely deep models and one traditional model were trained on the whole training set, and each model classified the test sample independently. The drug review test sample was then classified using the most confident classifier. The findings indicated that the 3W1DT and 3W3DT approaches outperformed standard and DL methods. Li et al. \cite{Li2020} introduced a novel padding approach that resulted in a more stable size of the input data sample and a rise in the amount of sentimental information present in each review. By using parallelisation, a DL-based SA model called lexicon was integrated with two-channel CNN–LSTM/BiLSTM family models. The experiments on various difficult datasets, such as the Stanford Sentiment Treebank, showed that the suggested approach is superior to a broad range of baseline techniques. Abid et al. \cite{Abid2019} designed a unified architecture for sentiment classification on Twitter that integrates an RNN model for extracting long-term dependencies with CNNs and GloVe as a word embedding approach. The experiments outperformed the baseline model on the Twitter corpora. Dang et al. \cite{Dang2021} proposed hybrid deep SA learning models that integrate LSTM networks, CNNs and SVMs. They were designed and evaluated on eight textual tweets and review datasets from various areas. On all datasets, the hybrid models outperformed the single models in SA. Fatemi and Safayani \cite{Fatemi2019} developed a generative approach based on NNs for joint sentiment topic modelling by changing the restricted Boltzmann machine structure and adding a layer corresponding to the sentiment of the text data. A contrastive divergence algorithm was used to manage and implement the proposed method. The new connected layer in the proposed method was a multinomial likelihood distribution layer, which could be used in textual sentiment classification or any other application.
\subsubsection{TEXTUAL DATASETS} \label{subsubsec:textual_datasets}
Researchers in SA might use their data or datasets that are publicly available. In order to be relevant to a certain problem, many researchers acquire new datasets relating to the subject. The main drawback is that the dataset must be labelled, which takes time. Furthermore, generating a large amount of data is not always simple. The availability and accessibility are the primary factors for selecting the datasets. Table \ref{tbl4:dataset_textual} highlights the most popular datasets used in textual SA.

\begin{table}
    \caption{Popular datasets for textual SA}
    \label{tbl4:dataset_textual}
    \begin{tabular}{p{100pt}p{350pt}}
    \hline
    \textbf{Datasets}   &\textbf{Links}  \\
    \hline
    IMDB    &\url{https://datasets.imdbws.com/} \\                             &\url{https://www.kaggle.com/datasets/lakshmi25npathi/imdb-dataset-of-50k-movie-reviews} \\
    &   \\
    \hline
    Stanford Sentiment Treebank &\url{https://nlp.stanford.edu/sentiment/treebank.html}\\
    Cornell Movie Reviews   &\url{https://www.cs.cornell.edu/people/pabo/movie-review-data/}    \\
    &   \\
    \hline
    \multirow{2}{100pt}{Thomson Reuters text research collection} &\url{https://dataverse.harvard.edu/dataset.xhtml?persistentId=doi:10.7910/DVN/IEJ2UX}  \\
    &   \\
    &\url{https://archive.ics.uci.edu/ml/datasets/reuters-21578+text+categorization+collection} \\
    &   \\
    \hline
    NYSK dataset	&\url{https://archive.ics.uci.edu/ml/datasets/NYSK#}    \\
    &   \\
    \hline
    ABC Australia News Corpus	&\url{https://live.european-language-grid.eu/catalogue/corpus/5162} \\
    &   \\
    \hline
    Sentiment labelled sentences dataset	&\url{www.kaggle.com/datasets/marklvl/sentiment-labelled-sentences-data-set}    \\
    &   \\
    \hline
    Sentiment 140	&\url{https://www.kaggle.com/datasets/kazanova/sentiment140}    \\
    &   \\
    \hline
    Tweets Airline	&\url{https://data.world/crowdflower/airline-twitter-sentiment} \\
    &   \\
    \hline
    \multirow{2}{100pt}{Tweets SemEval}  &\url{http://alt.qcri.org/semeval2016/task4/}   \\
    &\url{http://alt.qcri.org/semeval2015/task10/}  \\
    &\url{http://alt.qcri.org/semeval2017/task4/}   \\
    &   \\
    \hline
    \multirow{2}{100pt}{Book Reviews and Music Reviews}    &\url{https://data.world/dataquest/book-reviews}    \\
            &\url{https://kaggle.com/datasets/cakiki/muse-the-musical-sentiment-dataset}    \\
            &\url{https://webscope.sandbox.yahoo.com/catalog.php?datatype=r}    \\
    \hline
    \end{tabular}
\end{table}
\subsection{VISUAL SENTIMENT ANALYSIS} \label{subsec:visual_sentiment_analysis}
Popular social platforms, such as Flickr and Instagram, provide considerable amounts of visual data in the shape of images. VSA recognises and extracts sentiment and emotion from facial expressions or body movements. In addition to objects or entities, activities and locations, visual materials may include specific indicators that transmit attitude and emotion signals. For example, an image of a tasty meal, a gorgeous landscape or a lovely ceremony is likely to convey the publisher’s optimistic attitude. After condensing these sentimental experiences into meaningful labels, we may create computer vision tasks to identify appropriate mappings from the low-level visual input (e.g., raw pixels and motions) to the high-level emotional labels for classification, localisation, and summarisation functions. Throughout history, several studies have used image analysis to focus on specific cognitive and psychological applications, such as human face recognition detection and social marketing campaigns. This section describes the various image pre-processing operations, feature extraction methods and computational methodologies used in VSA.

\subsubsection{VISUAL PRE-PROCESSING} \label{subsubsec:visual_pre-processing}
Image pre-processing represents the fundamental processing of pictures. It aims to modify the image data by removing unwanted distortions or enhancing particular important visual properties for further processing and analysis. Some of the most often used pre-processing operations, namely, pixel brightness transformation (PBT)/brightness correction, geometric transformations, image filtering and segmentation, Fourier transform and image restoration, are discussed as follows.
\subsubsection*{3.2.1.1) PBT}
Brightness transformations improve the pixel brightness, and the pixel properties determine the transformation. In PBT, thevalue of an output pixel is exclusively determined by the value of the corresponding input pixel. Such operators include brightness and contrast changes, as well as colour correction and transformations. Contrast enhancement is an important component of image processing for human and computer vision. It is often utilised in medical image processing, voice recognition, texture synthesis and various other image or video processing applications. Brightness transformations may be categorised into brightness corrections and grayscale transformations. The most frequent procedures for PBT are gamma correction or power law transform, histogram equalisation and sigmoid stretching.
\begin{itemize}
    \item Gamma Correction. It is a nonlinear adjustment of the pixel value. Linear operations, such as scalar multiplication and addition/subtraction, are used on individual pixels in picture normalisation. Conversely, gamma correction uses a nonlinear operation on the source image pixels, which might change the image saturation.
    \item Histogram Equalisation. It is a popular contrast enhancement method that works on almost any picture format. It is a complex approach for adjusting an image’s dynamic range and contrast by redesigning its intensity histogram. Histogram modelling operators use nonlinear and non-monotonic transfer functions to map between pictures instead of linear and monotonic transfer functions.
    \item Sigmoid Stretching. A sigmoid function is a nonlinear activation function. The term ‘sigmoid’ comes from the S form of the function, which statisticians call a logistic function.
\end{itemize}
\subsubsection*{3.2.1.2) GEOMETRIC TRANSFORMATIONS}
A geometric transformation modifies the positions of pixels in an image while keeping the colours the same. It enables the reduction of geometric distortion that occurs during the acquisition of an image. The standard geometric transformation operations are rotation, shearing, translation, scaling, distortion (or un-distortion), Affine Transformation which combines the four transformations and Perspective Transformation, which alters the perspective of an image or video to gain a better understanding of the required information.

\subsubsection*{3.2.1.3) IMAGE FILTERING AND SEGMENTATION}
Filters are used to improve the visual qualities of images and/or extract important information, such as edges, corners and blobs. The kernel of a filter is a small array applied to each pixel and its neighbours in an image. Some of the basic filtering techniques are as follows.
\begin{itemize}
    \item Low-pass Filtering (Smoothing). It serves as the foundation for most smoothing methods. Smoothing a picture involves averaging nearby pixels to reduce the variance between pixel values.
    \item High-pass Filtering (i.e. Edge Detection, Sharpening). It makes an image look sharper. Unlike low-pass filters, high-pass filters enhance small details in the picture. It is similar to low-pass filtering except for using a different convolution kernel type.
    \item Directional Filtering. The first derivatives of a picture may be calculated using directional filtering, which is an edge detector. When the values of surrounding pixels shift considerably, the initial derivatives (or slopes) become particularly visible. Any spatial direction may be filtered with a directional filter.
    \item Laplacian Filtering. A Laplacian filter is an edge detector that measures the second derivatives of an image by measuring the rate at which the first derivatives change. It evaluates whether a change in the values of neighbouring pixels is caused by an edge or is part of a continuous development. Negative values are often included in Laplacian filter kernels as a cross pattern centred within the array. The corners may be zero or positive, and the centre may be positive or negative.
\end{itemize}
\subsubsection*{3.2.1.4) IMAGE SEGMENTATION}
Image segmentation is a common technique in digital image processing and analysis for dividing an image into segments or areas based on pixel attributes. It may separate the foreground and background pixels or cluster pixels based on colour or shape. The three types of image segmentation methods are as follows.
\begin{itemize}
    \item Non-contextual Thresholding. Thresholding is the easiest non-contextual segmentation approach. With a single threshold, this algorithm turns a greyscale or a coloured picture into a binary region map with a single threshold.
    \item Contextual Segmentation. Non-contextual thresholding combines pixels regardless of where they are on the image plane. Contextual segmentation can better identify various objects because it considers the closeness of pixels that belong to an object.
    \item Texture Segmentation. Many image analysis and computer vision applications rely on texture. It divides an image into areas with different textures that use comparable sets of pixels.
\end{itemize}
\subsubsection*{3.2.1.5) FOURIER TRANSFORM AND IMAGE RESTORATION}
Fourier transform is used to divide an image into sine and cosine components. The transformed image is in the Fourier or frequency domain, whereas the original image is in the spatial domain. Each point in the Fourier domain picture represents a frequency in the spatial domain image. Fourier transform is utilised in image analysis, filtering, reconstruction and compression.

Many researchers have used different types of pre-processing operations in their studies. Zhang et al. \cite{Zhang2016} proposed a method for detecting salient objects that immediately generates a limited set of detection windows for an input image. They created the silent objects’ proposals using a CNN model and proposed an optimised method for generating a limited set of detection windows from the noisy proposals. Navaz et al. \cite{Navaz2019} used single-image super-resolution and DL methods to increase the resolution of images in an emotion detection dataset. It is a significant pre-processing method that uses CNN to create high-quality images. Priya et al. \cite{TamilPriya2020} employed image augmentation techniques, including scaling, rotation and translation, to pre-process images in a dataset. During scaling, the object borders were trimmed, and rotation detected the item in any orientation.

\subsubsection{VISUAL FEATURE EXTRACTION} \label{subsubsec:visual_feature_eatrction}
The most difficult aspect of constructing a VSA system and designing a data analysis technique is selecting which data characteristics appropriately contain the information the system seeks to infer. Three important levels of semantics that are continually used for visual feature extraction are as follows.
\subsubsection*{3.2.2.1) LOW-LEVEL FEATURES}
These features explain diverse visual phenomena in a picture, most of which are related to the colour values of the image pixels in some manner. They often involve general characteristics, such as colour histogram (CH), histogram of oriented gradient (HOG) and GIST. Previous research on VSA \cite{Machajdik2010} revealed that low-level features, such as colours and textures, might be used to express an image’s emotional influence. Low-level features are divided into global and local features. Table \ref{tbl5:low-level_feature_extraction} highlights the most important low-level feature extraction methods used in the literature.

\begin{table}
    \small
    \caption{Description of the most important low-level feature extraction methods in VSA}
    \label{tbl5:low-level_feature_extraction}
    \begin{tabular}{p{30pt}p{70pt}p{350pt}}
    \hline
    \multicolumn{3}{c}{\textbf{Low-level/Global Features}}   \\
    \hline
    \textbf{Features}	&\textbf{Methods}	&\textbf{Description}   \\
    \hline
    \multirow{5}{20pt}{Colour}  &CH &CH was introduced by Swain et al. \cite{Swain1991}. It is an image retrieval method that visualises colour distribution within a picture. A one-dimensional feature vector can be used to represent a CH picture. CH methods are location-insensitive due to translation, rotation, and zoom invariance. They are appropriate for images that are challenging to colour-divide automatically or demand less pixel space.  \\
    \cline{2-3}
    &Colour Moments &It is a simple and effective colour characterisation method proposed by Stricker et al. \cite{Stricker1995}. Typical moments, which contain the standard deviation, mean, and skewness, can explain the average colour, colour variance and offset, and sufficiently define the colour distribution in pictures.  \\
    \cline{2-3}
    &Colour Coherence Vector    &It is a development of the CH method that was defined by Pass and Zabih \cite{Pass1996} for identifying the colour features of a picture. For each pair of pixels, the histogram is split into two halves. If a particular continuous area that is inhabited by certain pixels in the group is greater than a specified threshold, then the pixels are coherent, and this area is connected. If the threshold is not met, then the pixels are incoherent, and this region is disconnected.   \\
    \cline{2-3}
    &Colour Set (CS)    &CS, defined by Smith et al. \cite{Smith1996}, is considered a simplified version of the CH. It refers to a binary vector denoted by the symbol BM, an m-dimensional binary space, with each space axis corresponding to a single search m. When the colour m emerges, a CS, as a two-dimensional vector in the BM space, corresponds to the colour selection; c[m] = 1; otherwise, c[m] = 0. CSs may be formed instantly by specifying a specific threshold for the CH.    \\
    \cline{2-3}
    &Colour Correlogram &It was proposed by Huang \cite{Huang1998} as an enhancement over CH to describe the colour distribution of an image, which not only can display the percentage of pixels of a particular colour in an entire picture but also indicate the spatial connection between distinct colour pairings.    \\
    \hline
    \multirow{5}{20pt}{Texture} &Statistical Method: Grey-level Co-occurrence Matrix (GLCM) &It mainly uses the grey distribution to quantify textural features, including homogeneity, directionality and density of a picture. Using simple, robust, adaptable and easy-to-execute statistical techniques has many advantages. However, these approaches often require a large number of statistical computations. In terms of statistical analysis, GLCM,  defined by Mukherjee \cite{Mukherjee2016} and Savita et al. \cite{Savita2017} is the most often used method. \\
    \cline{2-3}
    &Model Method: Markov Random Field (MRF)    &It uses model parameters to represent texture characteristics. It specifies both the general regularity of the texture as well as the randomization of the texture location. MRF is a popular random field model. It is supported by a principle theory \cite{Vacha2011}, which emphasises that any pixel in the picture has a grey value that corresponds to the grey values of both the pixels surrounding it and the nonadjacent ones. \\
    \cline{2-3}
    &Structural Method: Texture Primitive (TP)  &It is a texture analysis technique based on the TP hypothesis presented by Tan and Constantinides \cite{Tan1989}. In TPs, a $2 \times 2$ window is the most basic component that can be used to describe the picture's texture. It is more appropriate to regular artificial texture; thus, the structural technique has substantial practical limitations. \\
    \cline{2-3}
    &Signal Method: Wavelet Transform (WT)  &WT is a representative signal approach introduced by Mallat \cite{Mallat1989} that is used for the feature extraction of image texture. It has a long history of being used in picture texture analysis. It decomposes the signal into a set of fundamental functions that can be created by changing the parent function.    \\
    \hline
    \multirow{5}{20pt}{Shape}   &Fourier Descriptor (FD)    &FD is a commonly used shape feature descriptor that describes the shape feature of an image via Fourier transformation of the object boundary; it has properties of low computational cost, clarity and flexibility to transition from coarse to fine description \cite{Shu2015}.    \\
    \cline{2-3}
    &Hough Transform (HT)   &It is a widely used technique proposed by Hough \cite{Hough1962} for distinguishing geometric forms that share certain characteristics from a picture and for recognising lines that appear at different angles.  \\
    \cline{2-3}
    &Invariant Moment (IMs) &Hu \cite{Ming-KueiHu1962} proposed IM based on algebraic invariants, which use the image’s grey distribution moments to characterise the distribution of grey features.    \\
    \cline{2-3}
    &Zernike Moment (ZM)    &ZM is an orthogonal complex moment in a polar coordinate space defined over the interior of the unit disc. ZM’s form characteristic is not susceptible to noise, and its value is not redundant because the kernel of the ZM is a set of orthogonal radial polynomials. These moments may characterise the shape and detail of an image.  \\
    \hline
    GIST descriptor &Global Image Descriptor    &It is a type of global  image feature critical for recognising an image's scenes. It is calculated by convoluting the filter with an image at different scales. As a result, an image's high- and low-frequency repeated gradient directions may be measured..   \\
    \hline
    \multicolumn{3}{c}{\textbf{Low-level/Local Features}}   \\
    \hline
    \multicolumn{2}{p{72pt}}{Local Binary Pattern (LBP)}  &The LBP function is a binary form of a texture analysis approach that compares the value of the centre pixel to its neighbours \cite{Ojala2000}. LBP recognises pixels in a picture by thresholding their neighbourhood and translating the result to a binary value. The LBP texture operator is frequently used due to its discriminating power and operational simplicity.    \\
    \hline
    \multicolumn{2}{p{72pt}}{HOG}   &HOG is a feature descriptor that calculates and counts the gradient direction histogram of an image's local region for computing objects in computer vision and image processing.   \\
    \hline
    \multicolumn{2}{p{72pt}}{Scale-invariant Feature Transform (SIFT)}  &SIFT identifies extreme points in a scale space and extracts position, scale and rotation invariance.   \\
    \hline
    \multicolumn{2}{p{72pt}}{Speeded-up Robust Feature (SURF)}  &SURF is a fast version of SIFT. It detects feature points using the determinant value of the Hessian matrix and speed up operations with integral graphs. \\
    \hline
    \multicolumn{2}{p{72pt}}{Bag of Visual Words (BOVW)}    &The BOVW model can be used for image classification or retrieval by treating image features as words. It is a vector of counted occurrences of a vocabulary of local image features.    \\
    \hline
    \end{tabular}
    \label{tbl:5}
\end{table}
\subsubsection*{3.2.2.2) MID-LEVEL FEATURES}
These features include greater semantic information, being easier to comprehend and having stronger emotional links. Many of the previous works on VSA have used the 1200D mid-level representation provided by Borth et al. \cite{Borth2013}. The advantage of using adjective noun pairs (ANPs) over nouns or adjectives is that a neutral noun may be transformed into a strong feeling ANP. Such paired notions are also more noticeable than adjectives alone. Table \ref{tbl6:mid-level_feature_extraction} highlights the most important low-level feature extraction methods used in the literature.

\begin{table}
    \caption{Description of the most important mid-level feature extraction methods in VSA}
    \label{tbl6:mid-level_feature_extraction}
    \begin{tabular}{p{100pt}p{350pt}}
    \hline
    \multicolumn{2}{c}{\textbf{Mid-level Features}}   \\
    \hline
    \textbf{Features}	&\textbf{Description}   \\
    \hline
    Visual Sentiment Ontology (VSO) &VSO was proposed by Borth et al. \cite{Borth2013}. One of its advantages is that it incorporates more classes in well-known visual ontologies, such as large-scale concept ontology for multimedia and ImageNet. It may be utilised in huge sentiment applications, such as microblog SA. VSO was first used to construct SentiBank, which is then used to predict the emotion of a picture. SentiBank is a library for trained concept detection, which offers a visual representation at the mid-level. Thus, it can provide better meaning than the low-level visual representation; experimentally, it is robust in describing images' emotional content. \\
    \hline
    Sentribute Mid-level Attribute  &Yuan et al. \cite{Yuan2013} developed a Sentribute method which uses sentimental scene attributes as mid-level features. Four mid-level attributes were selected: 1) material, 2) action, 3) surface aspect, and 4) spatial envelope. Finally, Sentribute provides 102 mid-level qualities that are easy to comprehend and ready to use. Sentribute is less strict than VSO because the four types of mid-level quality are specified individually, whereas VSO has become a prominent mid-level emotion feature.   \\
    \hline
    Object-based Visual Sentiment Concept Analysis &Although  SentiBank is beneficial for large-scale VSO, it faces two difficulties. That is, 1) object-based notions must be localised, and 2) ambiguity exists in the visual sentiment annotation. To address the issues above, Chen et al. \cite{Chen2014} developed object-based visual sentiment concept analysis. It is a hierarchical system that decomposes difficult issues into object localisation and sentiment-related idea modelling.  \\
    \hline
    Visual Sentiment Topic (VST) Model  &This VSO-based algorithm has two major issues with topic image SA. Firstly, A VSO-based method cannot  determine the ANP that most closely matches an image's primary sentiment. Second, microblog photos are significant for the same topic, but a VSO-based method cannot combine multi-image data to predict the correct sentiment. Cao et al. \cite{Cao2016} used multi-image information on the same subject to generate VST, focusing on the important VSO in a picture.   \\
    \hline
    \end{tabular}
\end{table}
\subsubsection*{3.2.2.3) HIGH-LEVEL FEATURES}
These features represent the semantic concepts depicted in the images at a high level. Such feature representation may be created using pre-trained classification algorithms or semantic embeddings \cite{Pilli2020}. Pre-trained CNN-based models (e.g. VGG16, VGG19 \cite{Simonyan2014}, Xception \cite{Chollet2017}, Inception V3 \cite{Szegedy2016} and Resnet \cite{He2016}) are state-of-the-art image classification models that extract high-level semantic information from photos. These models are pre-trained using imageNet, a massive visual database dedicated to the research on visual object identification.

\subsubsection{VISUAL CLASSIFICATION APPROACHES} \label{subsubsec:visual_calssification}
Modelling, recognising and utilising sentiments expressed by facial or body gestures or sentiments associated with visual multimedia are the key research topics in VSA. As a result, researchers are continually proposing, evaluating and comparing new methods to improve SA performance and solve the challenges in this sector. An overview of the categorization strategies frequently used for VSA is presented in this section. The majority of extant methods fall into three categories: ML, DL and transfer learning techniques. ML is the most often used method; it largely relies on manually developed features to extract and categorise visual information using ML techniques. To increase visual sentiment classification performance, DL algorithms extract the most important visual characteristics without using manually created features and overcome the limits of ML approaches. The visual sentiment classification issue may be addressed via transfer learning, a common technique in computer vision, because it enables the quick creation of correct models. Transfer learning is often demonstrated by using pre-trained models trained on a large ImageNet dataset to solve a problem similar to the one at hand. With the computational expense associated with training such models, models from the published literature (e.g. VGG, Inception and MobileNet) can be used. Numerous research has been conducted in this area. Afzal \cite{ShaikAfzal2021} developed an optimisation-based SVM model for automated VSA. The features of the input pictures are initially taken from the pre-trained ResNet-18’s weighed FC8 layer, where the relief method is used to analyse the modified weight. The SVM classifier was optimised using a hybrid optimisation methodology called the Holoentropy Life Choice Optimisation algorithm, which combines the advantages of life choice-based optimization and cross-entropy methods. The model was evaluated using the Emotion-6 and Abstract Art photo datasets, showing great performance. Instead of using SVMs, Jia et al. \cite{Jia2012} proposed a specific graphical model for classification using the same colour characteristics. A total of 23k digital photos were used, including images of paintings with descriptors such as pretty, casual, romantic and jaunty. Machajdik and Hanbury \cite{Machajdik2010} investigated the features that help with visual classification. The most essential visual features were selected based on emotional responses to colours and art. Emotional pictures were categorised by the writers’ eight emotional outputs (i.e. awe, anger, amusement, contentment, excitement, disgust, sad and fear). Amencherla and Varshney \cite{Amencherla2017} investigated the association between the visual content aspects of Instagram images and the psycholinguistic mood of their hashtag captions. Many colour attributes (e.g. hue, saturation and value), as well as colourfulness, hue and colour warmth, were used to predict visual emotion in thousands of photos. The findings support and clarify several psychological assumptions about the relationship between colour and mood/emotion, such as the relationship between colour and pleasure. Wu et al. \cite{Wu2020a} developed a VSA method that considers global and local data. Emotion was first deduced from the entire collection of photographs. Secondly, whether significant elements exist in a photograph was assessed. Sub-images were extracted from the entire image depending on the detection window of the notable elements. In addition, a CNN model was trained on the sub-images group. The final results were obtained by combining sentiment predictions from the photographs and sub-images. Yang et al. \cite{Yang2014} proposed a graphical model that displays the correlations between visual features and friends’ interactions (i.e. comments) on shared images. Features (e.g. saturation, saturation contrast, bright contrast, cool colour ratio, figure-ground colour difference, figure-ground area difference and background and foreground texture complexity) were utilised as visual characteristics. The model can distinguish between comments directly linked to an image's emotional expression and those unrelated.

Although most VSA projects incorporate many handcrafted visual elements, there is still no consensus on which visual characteristics are most relevant for a specific activity. Despite the benefits of previously investigated features, recent VSA research suggests that investigating mid-level representations as a link between low-level visual cues and sentiment orientation offers promise. Borth et al. \cite{Borth2013} generated a large VSO utilising psychological theories and web mining (SentiBank). ‘Beautiful flowers’ or ‘sad eyes’ are examples of ANP. The authors trained a collection of 1,200 visual concept detectors to determine the emotion associated with each picture. ANP detector outputs with a dimension of 1,200 may be used to train a sentiment classifier. They collected adjectives and nouns from YouTube and Flickr videos and picture tags. The Plutchik Wheel of Emotion, a well-known psychological model of human emotions, was utilised to find these pictures and videos. The researchers showed a large tagged picture collection of half a million Flickr photographs linked to 1,200 ANPs. The results revealed that SentiBank concepts outperform text-based approaches in tweet sentiment prediction. Li et al. \cite{Li2018} developed an approach for completely using the textual sentiment information included in ANPs. They used the one-dimensional visual feature Vector and the ANPs' textual sentiment scores to infer the overall sentiment value of an image. Yuan et al. \cite{Yuan2013} initially used scene-based attributes to generate mid-level features and then built a binary sentiment classifier on top of them. They indicated that adding a step for facial expression recognition increases sentiment prediction when deployed to images featuring faces.

Due to their accessibility and adaptability, VSO and SentiBank are commonly utilised in predicting viewer emotions for photos and animated GIFs \cite{Jou2014}. Although the majority of current mid-level representation-based methods can construct an ontology of sentiment, they disregard the distinctions and connections among ontology categories. A recent trend in SA and opinion mining for visual information is based on recent achievements in computer vision with DL. Desai et al. \cite{Desai2020} intended to solve the issues of VSA by using a DL CNN and affective regions technique to provide intelligible sentiment reports with high accuracy. Chen et al. \cite{Chen2020} presented a unique active learning architecture that requires minimal labelled training data. Firstly, a new branch called the ‘texture module’ was added to the CNN. Computing the inner products of feature maps from various convolutional blocks in this branch led to generating the emotional vector, which was then used to distinguish effective pictures. Secondly, a query strategy was constructed using the standard CNN and the texture module classification scores. The samples produced by the query were then utilised to train the model. Extensive testing on four public affective datasets showed that the method works well for VSA with few labelled training samples. Song et al. \cite{Song2018} introduced sentiment networks with visual attention, a unique architecture that combines visual attention into the well-established CNN sentiment classification framework through end-to-end training. Wu et al. \cite{Wu2020} suggested a multitask learning technique for visual attribute identification. The semantic gap between visual characteristics and subjective attributes may be closed by adding sentiment supervision to the attributes. Then, a multi-attention model was used to identify and localise numerous relevant local areas based on expected attributes. The classifier constructed on top of these areas considerably improves the prediction of the visual sentiment. The strategy is better than earlier methods in experiments. Cetinic et al. \cite{Cetinic2019} used CNNs to estimate scores associated with three subjective aspects of human perception: image aesthetic evaluation, mood evoked by the image and image memorability. For each topic, many unique CNN models were trained on various natural image datasets, and the model with the highest performance was selected based on qualitative results and compared to existing subjective artwork ratings. Yang et al. \cite{Yang2018} described a strategy for locating emotional zones using a deep framework. They used a regular technique to produce N object propositions from a query image and scored them according to abjectness. A pre-trained and fine-tuned CNN model was then used to generate the sentiment score for each proposition. On the basis of both scores, the top K locations were selected from a pool of N applicants. Finally, deep features were taken from the entire image, as well as selected regions and a sentiment label was projected. Numerous large-scale datasets have demonstrated that the approach can identify emotional local regions and achieve state-of-the-art results. Ou et al. \cite{Ou2021} proposed a multilevel context pyramid network to improve the classification performance of VSA. To begin, they utilised Resnet101 to obtain multilevel emotional representation. Subsequently, multiscale adaptive context modules were used to determine the degree of sentiment connection between diverse sections of varying sizes in the image. Finally, many layers of contextual information were combined to create a multi-cue emotional feature for categorisation of image sentiment. Extensive testing on typical visual sentiment datasets showed the efficiency of the strategy. Yadav et al. \cite{Yadav2020} proposed a residual attention-based DL network for VSA to use CNN to learn the spatial hierarchies of visual features. A residual attention model was employed to focus on the image’s critical sentiment-rich local areas. This study made a significant addition by conducting a comprehensive investigation of seven major CNN-based architectures, including VGG-16, VGG-19, ResNet-50, Inception-Resnet-V2\&V3, Xception and NASNet. The impact of tuning on these CNN variations was presented in the VSA field.
\subsubsection{VISUAL DATASETS} \label{subsubsec:visual_dataset}
SA datasets may be collected from various sources. The traditional method of acquiring a set of human-labelled data is to apply large-scale surveys on a massive group of people. Conversely, in SA, a large amount of opinion information can be obtained by leveraging popular social networks (e.g. Instagram, Flickr, Twitter and Facebook) along with webpages devoted solely to gathering business and product reviews. Table \ref{tbl:7} provides an overview of the most significant datasets in VSA.

\begin{table}
    \caption{Overview of the most significant datasets in VSA}
    \begin{tabular}{p{100pt}p{350pt}}
    \hline
    \textbf{Datasets}   &\textbf{Links}  \\
    \hline
    International Affective Picture System (IAPS)    &\url{https://csea.phhp.ufl.edu/media.html} \\
    \hline
    Emotional Category Data from IAPS    &\url{https://link.springer.com/article/10.3758/BF03192732}\\
    \hline
    Affective Image Classification Dataset &\url{http://www.imageemotion.org/}    \\
    \hline
    Flickr Sentiments   &\url{http://www.l3s.de/_minack/flickr-sentiment/}  \\
    \hline
    Geneva Affective Picture Database   &\url{https://www.unige.ch/cisa/index.php/download file/view/288/296/} \\
    \hline
    VSO &\url{https://visual-sentiment-ontology.appspot.com/} \\
    \hline
    Emotion6 	&\url{http://chenlab.ece.cornell.edu/downloads.html}  \\
    \hline
    Twitter Images	&\url{https://www.cs.rochester.edu/u/qyou/DeepSent/deepsentiment.html}    \\
    \hline
    Cross Sentiment	&\url{http://mm.doshisha.ac.jp/senti/CrossSentiment.html}  \\
    \hline
    T4SA 	&\url{http://www.t4sa.it/} \\
    \hline
    Flickr 8k Dataset	&\url{https://shorturl.at/iuBLP}   \\
    \hline
    POM Movie Review Dataset	&\url{http://multicomp.cs.cmu.edu/resources/pom-dataset/}  \\
    \hline
    Flickr Image Dataset for VSO &\url{https://www.ee.columbia.edu/ln/dvmm/vso/download/flickr_dataset.html}    \\
    \hline
    Twitter Image Dataset for VSO &\url{https://www.ee.columbia.edu/ln/dvmm/vso/download/twitter_dataset.html}   \\
    \hline
    ICT YouTube Opinion Dataset &\url{http://multicomp.cs.cmu.edu/resources/youtube-dataset-2/}    \\
    \hline
    \end{tabular}
    \label{tbl:7}
\end{table}
\subsection{VISUAL-TEXTUAL SENTIMENT ANALYSIS} \label{subsec:visual-textual_sentiment}
People express their views on social networking platforms like Twitter, Facebook,  Flicker and Instagram. These user-generated materials have become more diverse in terms of substance and structure, with users increasingly posting text-embedded pictures called ‘image–text postings’. Different from traditional text-only articles, these blogs are more informative because they provide visual materials in addition to text. SA attempts to automatically determine the underlying sentiment of the postings by combining textual and visual materials that may contribute more to understanding user feelings and behaviour. The combination of several social contents, their related characteristics or decisions for performing an analytical task is known as multimodal fusion. Numerous definitions of information fusion have been presented in the literature based on \cite{Bostrom2007}. Combining many modalities to achieve good performance in various applications is better than using a single modality. This method has attracted increasing attention from researchers across a range of disciplines due to its potential for an infinite number of applications, including SA, emotion recognition, human tracking, semantic concept detection, image segmentation, event detection and video classification. Information fusion can improve sentiment performance by assuming that the heterogeneity of many information sources allows the cross-correction of certain errors, which leads to enhanced results. Utilising various information sources could complement knowledge and boost overall decision-making accuracy. To establish the most appropriate fusion approach, the following fundamental questions should be addressed: What is the suitable level to implement the fusion strategy, and how can the information be fused?

For the task of visual–textual sentiment classification, three levels of fusion, namely, early- or feature-level fusion, intermediate- or joint-level fusion and late- or decision-level fusion, can be defined based on the type of available information in a certain field. Table \ref{tbl8:most_popular_fusion} shows an overview of these fusion techniques along with their advantages and disadvantages.

\begin{table}
    \caption{Most popular fusion techniques in visual–textual SA}
    \label{tbl8:most_popular_fusion}
    \begin{tabular}{p{50pt}p{90pt}p{145pt}p{145pt}}
    \hline
    \textbf{Fusion}  &\textbf{Description}  &\textbf{Advantages}  &\textbf{Disadvantages}   \\
    \hline
    Early- or feature-level fusion  &It is the process of combining the features extracted from several sources into a single feature vector, which is then fed into an ML algorithm.  &\begin{itemize}[leftmargin=*]
                    \item Can make early use of the association between several features obtained from various modalities, eventually leading to better performance.
                    \item Only needs a single learning phase to be performed on the merged feature vector.
                    \item Does not always need extensive design work
                    \item Does not need the training of several models
                \end{itemize}  &\begin{itemize}[leftmargin=*]
                                    \item Cannot capture the complementary nature of the modalities
                                    \item Generates high-dimensional feature vector that may even contain redundancies
                                    \item Requires all features to have the same format before initialising the fusion process
                                    \item Cannot capture the time-synchronicity of several sources due to the fact that the characteristics from separate but tightly connected modalities might be retrieved at different times
                                    \item Learning the cross-correlation among the diverse features becomes challenging as the number of modalities increases.
                                \end{itemize}   \\
    \hline
    Intermediate- or joint-level fusion &It is the process of combining learned feature representation from intermediate layers of NNs with features from other modalities as input to a final model. Most intermediate fusion models combine units with connections from various modality-specific channels using a shared representation layer.   &\begin{itemize}[leftmargin=*]
                \item Can simulate interactions between features from several modalities
                \item Can learn more enhanced feature representation from each modality
                \item Can combine input at various abstraction levels
                \item Does not require the training of several models
            \end{itemize}   &\begin{itemize}[leftmargin=*]
                                \item It might result in overfitting, wherein the network fails to simulate the relationship between both modalities accurately. Therefore, a careful design is required
                                \item Cannot perform SA effectively when a portion of multimodal information is missing
                            \end{itemize}   \\
    \hline
    Late- or decision-level fusion  &It is the process of integrating the results of numerous sentiment classifiers that have been trained independently for each modality.   &\begin{itemize}[leftmargin=*]
        \item Can predict even when not all modalities exist
        \item Having a large quantity of training data is not required
        \item No need to convert data to the same format
        \item Great flexibility since the best techniques are used to analyze each modality individually
        \item Scalable in terms of the modalities employed in the fusion process (i.e. gentle up-gradation or degradation).
                    \end{itemize}   &\begin{itemize}[leftmargin=*]
                                        \item Cannot capture the relationship between several modalities
                                        \item Ignores the modalities’ low-level interaction
                                        \item Ensemble all classifiers effectively is difficult
                                        \item Probability of missing local interactions between modalities
                                    \end{itemize}   \\
    \hline
    \end{tabular}
\end{table}
\subsubsection{MULTIMODAL FUSION METHODS} \label{subsubsec:multimodal_fusion}
This section discusses many fusion strategies that have been employed to address various visual-textual SA challenges. The fusion approaches are classified as follows: rule-, classification-, attention- and bilinear pooling-based approaches. The optimal strategy for a specific application depends on the nature of the issue, the characteristics of the media and the available parameters.

\subsubsection*{3.3.1.1) RULE-BASED FUSION METHODS}
A number of essential requirements for combining multimodal data are included in rule-based fusion techniques. These methods include linear weighted fusion (sum and product), MIN, MAX, OR, AND, majority voting and custom-defined rules. These concepts were conceptually presented by Kittler et al. \cite{Kittler1998}. The linear weighted fusion method is considered one of the most fundamental and extensively employed techniques. It uses sum or product operators to fuse the information received from distinct modalities or decisions derived from a classifier. To combine the data, normalised weights should be allocated to individual modalities. Weight normalisation approaches published in the literature include sigmoid function, tanh estimators, z score, min-max, and decimal scaling. Each of these techniques has advantages and disadvantages. When the matching scores for the different modalities are easily acquired, the z-scoring, decimal scaling, and min-max methods are favoured. However, these methods are susceptible to outliers. Although the \textit{tanh} normalization approach is reliable and effective, it does need parameter estimation via training. Linear weighted fusion technique is less expensive computationally than other techniques. However, for effective execution, the weights should be appropriately normalized. Majority voting fusion is employed depending on the decision reached by most of the classifiers. Custom-defined rules are explicitly established for an application. They are developed based on data collected from several modalities and the intended ultimate consequence to obtain optimum judgements similar to the work in Kumar et al. \cite{Kumar2020}.
\subsubsection*{3.3.1.2) CLASSIFICATION-BASED FUSION METHODS}
This approach uses various classification methods to categorise multimodal data into predefined groups. SVMs, Bayesian inference, Dempster–Shafer theory, dynamic BN (DBN), NNs, and ME models are all examples of approaches included in this category. SVM is possibly the most extensively utilised approach for supervised learning on data categorisation jobs. The approach classifies the incoming data vectors into pre-set learnt classes, thereby resolving the pattern classification issue in the context of multimodal fusion. The Bayesian inference fusion approach combines multimodal data according to the probability theory’s criteria. This approach integrates features from many modalities or classifiers to produce a joint likelihood inference. The DBN is a graph network in which the nodes represent distinct modalities, where the connections show the probabilistic dependencies between these nodes. The benefit of this network over other approaches is that it facilitates a simple combination of the temporal dynamics of multimodal data. The second approach that is often employed is NNs. A common NN model comprises input, hidden and output layers. The network may be fed with data in the form of multimodal characteristics or classification choices from several classifiers. Results contain a fusion of the data being considered. The activation functions required to generate the predicted output are supplied by the hidden layer of neurons, and the number of neurons and hidden layers is selected to provide the required level of accuracy. The technique is most often used for decision-level fusion.

\subsubsection*{3.3.1.3) ATTENTION-BASED FUSION}
The fusion process mostly uses the attention mechanism like in Zhang et al. \cite{Zhang2020} work, which refers to the weighted sum of a collection of vectors with dynamically generated scalar weights at each step by a small ‘attention’ model. Numerous glimpses, sometimes known as output heads, are frequently utilised in order to generate several dynamic weight sets for summing, which could preserve extra details by combining the outcomes of each glimpse.

When attention mechanisms are performed on an image similar to the work of Zhang et al. \cite{Zhang2021}, the feature vectors that are related to distinct regions are given different weights to construct an attended image vector, which is then used to associate the emotional image components with their related text description.

In contrast to image attention mechanisms, Cao et al. \cite{Cao2020} applied the co-attention mechanisms, which utilise a symmetric attention framework to create attended text and image feature vectors, whereas Xu et al. \cite{Xu2021} used the dual attention network (DAN) which is identical to the parallel co-attention in predicting the attention distribution for pictures and languages concurrently. Such attention models work based on modalities' characteristics and memory vectors. Different from co-attention, memory vectors may be repeatedly modified at every reasoning level by utilizing repetitive DAN structures.

Conversely, Xu et al. \cite{Xu2019} presented an alternating co-attention mechanism that first produces an attended image vector based on linguistic featuresand then produces an attended language vector based on the attended picture vector.

Multi-head attention was also proposed by Yang et al. \cite{Yang2021} and Zhang et al. \cite{Zhang2022} to define the correlation between the image and text contents. Furthermore, gated multimodal fusion is considered a different type of attention because it uses gating to assign weights to visual and textual elements. The weighted sum of the feature vectors for both visual and textual features may be computed on the basis of the dimension-specific scalar weights that were generated automatically using a gated process. Then, multimodal sentiment classification may be performed using these representations, like in the works of Huang et al. \cite{Huang2020} and Arevalo et al. \cite{Arevalo2017}.
\subsubsection*{3.3.1.4) BILINEAR POOLING-BASED FUSION}
Bilinear pooling, also called order pooling \cite{Tenenbaum2000}, is a technique for integrating visual and textual information into a single representation space by calculating the outer product between the feature vectors of these modalities, allowing all the elements in both vectors to interact in a multiplicative manner. Bilinear pooling has an $n^2$-dimension representation generated by linearising the matrix  into a vector created using the outer product. This characteristic makes it more definitive than other straightforward operations, represented by the weighted sum, concatenation, or element-wise multiplication that generates $n$ or $2n$-dimension representations (assuming each vector has n elements). Typically, a two-dimensional weight matrix is used to linearly convert bilinear representations to output vectors, which is analogous to fusing two input feature vectors using only a three-dimensional tensor function. When computing an outer product, every feature vector may be expanded by one to retain single-modal input features \cite{Zadeh2017}. Conversely, bilinear pooling needs the decomposition of weight tensors to permit the related model to be trained appropriately and effectively due to its enormous dimensionality (usually on the scale of hundreds of thousands to millions of dimensions) \cite{Ye2019}. Bilinear pooling can be combined with attention mechanisms \cite{Zhou2020,You2016a,Zhu2022} to obtain optimum correlation between visual and textual contents.

\subsubsection{VISUAL–TEXTUAL SENTIMENT CLASSIFICATION LITERATURE}
Although SA upon single modality data has seen significant success in recent years, it cannot successfully manage the variety of information in social media data. Multimodal data, which combines images and text, has become increasingly common on social media websites. Such vast amounts of multimodal input can assist in comprehending how individuals feel or think about specific situations or topics. As a result, many multimodal sentiment categorisation approaches have been proposed to incorporate diverse modalities. These approaches are classified into three distinct categories: early/feature fusion, intermediate/joint fusion, and late/decision fusion.

In the early fusion strategy, a unified feature vector is first constructed, and then the features retrieved from the input data are fed into an ML classifier. Xu et al. \cite{Xu2018} characterised the relationship between text and picture using a co-memory attentional method. Although the reciprocal influence of text and picture was considered, a coarse-grained attention technique has been implemented, which may make it difficult to retrieve adequate information. In contrast, the co-memory network merely uses a weighted visual/textual vector like a reference to discover attention weights for visual/textual representation. It may be considered a coarse-grained attention mechanism that can lead to information loss because attending several items with a single attention vector might obscure the characteristics of each attended content. Zhang et al. \cite{Zhang2021} suggested a unique cross-modal semantic content correlation strategy using hierarchical networks and deep matching as foundations. The proposed joint attention network can learn the content association between an image and its caption, exported as a pair of images and texts. A class-aware sentence representation (CASR) network further processed the caption and a class dictionary. A fully connected layer was used to concatenate the CASR outputs and convert them into a single class-aware vector, which was finally integrated with the image–text pair as inputs to an inner-class dependency LSTM for extracting the cross-modal nonlinear associations for predicting the sentiment of the model. Arevalo et al. \cite{Arevalo2017} introduced a unique multimodal learning model based on gated NNs. The gated multimodal units (GMU) model was meant to be utilised as an intrinsic unit in an NN architecture to construct an intermediate representation using a set of inputs obtained from different modalities. The GMU learns to determine how modalities affect the unit's activation using multiplicative gates. Baecchi et al. \cite{Baecchi2016} investigated applying multimodal feature learning techniques, such as SG and denoising autoencoders, for SA of micro-blogging material, such as short Twitter messages, including text and an image. A unique architecture that combines these NNs was presented, and its efficiency and classification accuracy on numerous typical Twitter datasets were demonstrated. Peng et al. \cite{Peng2022} introduced a CMCN (cross-modal complementary network) for MSC, including hierarchical fusion. The CMCN is designed as a hierarchical network consisting of three key modules: cross-modal hierarchical fusion, feature extraction from texts and images, and feature attention from an image–text association generator. In this manner, a CMCN may assist in limiting the risk of incorporating unrelated modal features, and it outperforms existing approaches. Ji et al. \cite{Ji2019} presented a unique bilayer multimodal hypergraph learning (bi-MHG) for enabling accurate sentiment analysis in multimodal tweets. A two-layer framework for the suggested bi-MHG model (tweet-level hypergraph and feature-level hypergraph) was developed. In a bilayer learning system, multimodal characteristics are shared between the two layers. Bi-MHG specifically designs the importance of modality rather than intuitively weighting multimodal information as in conventional multimodal hypergraph learning. Finally, layered alternating optimisation was presented for parameter learning. Zhang et al. \cite{Zhang2020} developed a novel multimodal sentiment model that can remove the noise from textual input whilst extracting important visual characteristics. Then, they used the attention mechanism for feature fusion, where the textual and visual contents learn the intrinsic characteristics from each other by symmetry. Fusion characteristics were then used to classify the sentiments. Table \ref{tbl10} provides a summary of the literature that employs the early fusion methodology.

\begin{table}
    \centering
    \caption{Summary of the literature using early fusion}
    \label{tbl10}
    \begin{tabular}{p{70pt}p{100pt}p{60pt}p{80pt}p{95pt}}
        \hline
        \textbf{Author}  &\textbf{Features}   &\textbf{Classification Approach} &\textbf{Dataset}    &\textbf{Performance Metric} \\
        \hline
        \multirow{2}{70pt}{Xu et al. \cite{Xu2018}} & \textbf{Visual:} pre-trained CNN &\multirow{2}{60pt}{Soft-max Classifier}  &MVSA-single &Accuracy: 70.51\% \\
        &\textbf{Textual:} GloVe &  &MVSA-multiple &Accuracy: 68.92\% \\
        \hline
        \multirow{2}{70pt}{Ke Zhang et al. \cite{Zhang2021}} &\textbf{Visual:} VGG-19 &\multirow{3}{60pt}{IDLSTM} &Flicker & Accuracy: 84.2\% \\
        &\textbf{Textual:} GloVe   &  &Getty-image & Accuracy: 80.6\% \\
        &\textbf{Class Dictionary:} GloVe &  &Twitter & Accuracy: 86.3\% \\
        \hline
        \multirow{3}{70pt}{Arevalo et al \cite{Arevalo2017}} &\textbf{Visual:} VGG, end2end CNN  &Early / LR &\multirow{3}{80pt}{Multimodal IMDB}    &F1-score: 60\% \\
        &\multirow{2}{100pt}{\textbf{Textual:} N-gram, Word2vec} &Joint / MLP    &   &F1-score: 61.7\%   \\
        &   &Late / Average &   &F1-score: 60.4\%   \\
        \hline
        \multirow{2}{70pt}{Baecchi et al. \cite{Baecchi2016}} &\textbf{Visual:} Denoising autoencoder &\multirow{2}{100pt}{LR}  &\multirow{2}{80pt}{SentiBank Twitter}  &Accuracy: 79\% \\
        &\textbf{Textual:} Word2vec &  &  &  \\
        \hline
        \multirow{3}{70pt}{Peng et al. \cite{Peng2022}} &\textbf{Visual:} VGG   &\multirow{3}{60pt}{Soft-max}    &MVSA-single & Accuracy: 73.61\% \\
        &\multirow{2}{100pt}{\textbf{Textual:} Bert} &  & MVSA-multiple & Accuracy: 70.45\% \\
        &  &  & Multi-ZOL & Accuracy: 74.28\% \\
        \hline
        \multirow{3}{70pt}{Ji et al. \cite{Ji2019}} &\textbf{Textual:} BOVW using SentiBank &\multirow{3}{60pt}{Bi-Layer Multimodal Hypergraph Learning}   &\multirow{3}{80pt}{Cross modality} &\multirow{3}{*}{Accuracy: 90.0\%} \\
        &\textbf{Visual:} Bag of Textual Words &  &  &  \\
        &\textbf{Emoticon:} bag of emoticon words &  &  &  \\
        \hline
        \multirow{2}{70pt}{Zhang et al. \cite{Zhang2020}} &\textbf{Textual:} Denoising Autoencoder &\multirow{2}{100pt}{Soft-max}    &MVSA-single &Accuracy: 71.44\% \\
        &\textbf{Visual:} Attention-based Variational Autoencoder &  &MVSA-multiple &Accuracy: 69.62\% \\
        \hline
    \end{tabular}
\end{table}

According to the concept of intermediate fusion, the integration process occurs at the intermediate levels of the network. A shared representation layer connects units from distinct paths specific to multiple modalities. Zhou et al. \cite{Zhou2020} presented a cross-modality hierarchical interaction model for SA. The method handles noise and joint comprehension concerns by extracting semantic (bilinear attention mechanism) and sentiment interactions (multimodal CNN) between picture and text in a hierarchical approach. A hierarchical attention technique was initially used to collect semantic connections and filter information in one mode using another. Then, a multimodal CNN was used to fully leverage cross-modal sentiment connection, resulting in superior visual—textual representation. Further research on a transfer learning technique was conducted to decrease the effect of noise on actual social data. Jiang et al. \cite{Jiang2020} suggested a fusion extraction network model for multimodal SA. To begin, the model employs an approach to fuse the information interactively, thus acquiring a dynamic understanding of the visual-specific textual as well as the textual-specific visual representations. Then, for the particular textual and visual representations, an information extraction technique is utilised to extract meaningful information and filter out unnecessary components. Zhao et al. \cite{Zhao2019} presented a novel image–text consistency-driven multimodal SA technique to investigate the picture-text interaction, which will then be followed by a method for multimodal adaptive SA. The standard SentiBank technique was utilised to describe the visual cues by extracting the mid-level visual features with the corporation of additional characteristics, such as textual, low-level visual and social characteristics, to construct an ML SA approach utilising the VSO as a benchmark dataset. Guo et al. \cite{Guo2021} proposed an end-to-end news sentiment recognition method using Layout-Driven Multimodal Attention Network (LD-MAN). Instead of modelling text and pictures separately, LD-MAN aligns images with text using an online news layout. LD-MAN represents picture positions and quantifies image-text contextual interactions using a series of distance-based coefficients. LD-MAN then employs multimodal attention to understand the emotive representations of the articles from the aligned text and visuals. Huang et al. \cite{Huang2020} suggested a unique method called attention-based modality-gated networks (AMGN) to utilise the interaction across both modalities and derive the distinguishing characteristics for multimodal SA. A visual semantic attention model was presented to acquire certain visual characteristics of each word. A modality-gated LSTM Was suggested to obtain multimodal characteristics by selecting the modality that adaptively provides the most sentiment data. An automated semantic self-attention model was then suggested to focus more on the distinguishing characteristics for sentiment classification.

You et al. \cite{You2016} suggested a cross-modality consistent regression (CCR) model that incorporates visual–textual SA methods. Then, a CNN was used to extract the visual characteristics, and the textual features were learned using a distributed paragraph vector model trained on the titles and descriptions of images. Several fusion approaches, such as early fusion, late fusion and CCR, were used to conduct integrated visual-textual SA. By enforcing consistent constraints across related modalities, the CCR model was developed in order to derive the ultimate sentiment classifier. Duong et al. \cite{Duong2017} developed basic approaches (i.e. common feature space Fusion using an auxiliary learning task and joint fusion using a pooling layer) for classifying social media material incorporating information from several modalities. Xu et al. \cite{Xu2019a} suggested a unique hierarchical deep fusion model to study cross-modal connections among pictures, words and social ties for more effective SA. Specifically, three-level hierarchical LSTMs were used to discover the image-text relationships. A weighted relation network was also employed to exploit the connection information effectively, with each node in a distributed vector. The obtained image–text features and node embeddings were fused using a multilayer perceptron (MLP) to capture nonlinear cross-modal correlations further. Xu et al. \cite{Xu2021} suggested an attention-based heterogeneous relational model incorporating rich social information into multimodal SA to increase performance. A progressive dual attention module was designed to record image–text interactions and subsequently train the combined visual and textual representation from the standpoint of content information. Here, a channel attention schema was provided to emphasise semantically rich picture channels, and a region attention schema was presented to identify emotional areas based on the attended channels. The researchers next built a heterogeneous relation network and adapted the graph convolutional network to integrate material information from social contexts as supplements to create improved representations of social pictures. Zhu et al. \cite{Zhu2022} introduced a novel image–text interaction network to examine the association between expressive picture areas and text for multimodal SA. Specifically, a cross-modal alignment module was used to capture area word correspondence, which was then fused by an adaptive cross-modal gating module. They also incorporated individual modal contextual feature representations to provide a more accurate prediction. Tashu et al. \cite{Tashu2021} demonstrated a multimodal emotion detection architecture that leverages feature-level (sequential co-attention) and modality attention to categorise emotion in art. The proposed architecture allows the model to learn informative and refined representations for feature extraction and modality fusion. The proposed approach performed well on the WikiArt emotion dataset. Ortis et al. \cite{Ortis2021} suggested extracting and applying an Objective Text description of pictures instead of the traditional Subjective Text given by users to determine the sentiment related to social pictures. Objective text is extracted from photos using modern DL architectures to categorise objects and scenes and conduct image captioning. The objective text characteristics are merged with visual information in a canonical correlation analysis embedding space. Then, SVM was used to infer the sentiment polarity. The research demonstrated that using text retrieved from photos instead of text given by users enhances classifier performance. Cai et al. \cite{Cai2015} proposed a sentiment prediction system based on CNN for multimedia data. They employed two independent CNNs to obtain linguistic and image properties and then combined their outputs into another CNN to investigate the text–image relationships completely. You et al. \cite{You2016a} offered a novel approach for robust SA that combines text and visual data. Unlike earlier research, the authors claimed that visual and textual data should be addressed structurally. They initially constructed a semantic tree structure using sentence parsing to perform a correct analysis. The system then learned a strong visual–textual semantic representation by utilising an attention mechanism in conjunction with LSTM and an additional semantic learning task.

Yang et al. \cite{Yang2021a} developed an innovative multimodal emotion analysis model built on the multi-view attentional network for determining the deep semantic features of images and texts. The framework consists of three steps: interactive learning, feature fusion and mapping, which use stacking-pooling and multilayer perceptrons to fuse the various features profoundly. The TumEmo image–text emotion dataset was created as part of this study. Yadav and Vishwakarma \cite{Yadav2022} presented a deep multilevel attentive network to analyse visual and textual correlations. A bi-attentive visual map was created using spatial and channel dimensions to increase CNN's capability for representation. Then, the association between the bi-attentive visual features and word semantics was established by employing semantic attention. Lastly, sentiment-rich multimodal characteristics were effectively harvested using self-attention. Miao et al. \cite{Miao2018} determined the essential areas of an image based on its word representation. The picture portions were considered memory cells, and the attention process was used to retrieve them. Then, CNN was used to blend visual and textual information more efficiently using the IMDB dataset. Zhu et al. \cite{Zhu2019} suggested that visual and textual information's influence on SA should be differentiated. By including a mechanism for cross-modal attention and semantic embedding of knowledge using a bidirectional RNN, the model develops a strong joint visual–textual representation, making it better than the current state-of-the-art methods. MultiSentiNet is a deep semantic network proposed for multimodal SA by Xu and Mao \cite{Xu2017}. In order to extract meaningful semantic information from pictures, the authors used detectors for salient objects and scenes. Then, a visual feature-guided attention LSTM model was developed to capture important words to understand the feeling conveyed by an entire tweet and merge the representations of those significant words using visual semantic characteristics, objects and scenes. Experiments showed the efficiency of the MultiSentiNet model. Yang et al. \cite{Yang2021} developed a multichannel graph NNs with sentiment awareness for image–text sentiment detection. To capture hidden representations, they initially encoded multiple modalities. Then, multichannel graph NNs were used to learn multimodal representations using global dataset properties. Finally, multimodal in-depth fusion was used to estimate the emotion of image–text pairings. The various Syncretic co-attention networks were presented by Cao et al. \cite{Cao2020} to identify multilevel matching interactions across multimodal data and consider each modality's distinctive information for unified complimentary sentiment analysis. A multilevel co-attention module was constructed to study localised correspondences between picture regions and text words, as well as holistic correspondences across global visual data and context-based textual meaning. Then, all single-modal characteristics may be combined from multiple levels. The suggested VSCN additionally incorporates unique information about each modality concurrently and combines them into an end-to-end system for SA. Ye et al. \cite{Ye2019} developed visual–textual sentiment based on product reviews. Their study has two significant contributions. Firstly, a new Product Reviews dataset was developed instead of scraping data from Flickr or Twitter with positive and negative labels. Secondly, a deep Tucker fusion approach was proposed for visual–textual SA using Tucker decomposition and bilinear pooling operation to integrate deep visual and textual representations. Xu et al. \cite{Xu2019} introduced a unique bidirectional multilevel attention model to jointly classify visual and textual sentiments using complementary and complete information from image and text modalities. Liao et al. \cite{Liao2022} presented an image–text interactive graph NN for SA. The graph's node features started with text and picture features and were updated by the graph's attention mechanism. Finally, it would be merged with an image–text aggregation layer to achieve sentiment classification. Truong et al. \cite{Truong2019} proposed a visual–textual attention network based on the observation that images can support the text. The proposed model depends on visual data to highlight the important texts of the whole document. The summary of the literature using joint fusion is illustrated in Table \ref{tbl11}.

\begin{longtable}{p{70pt}p{100pt}p{60pt}p{80pt}p{95pt}}
    \caption{Summary of the literature using joint fusion} \label{tbl11}\\
    \hline
    \textbf{Author}  &\textbf{Features}   &\textbf{Classification Approach} &\textbf{Dataset}    &\textbf{Performance Metric} \\
    \hline
    \multirow{2}{70pt}{Zhou et al. \cite{Zhou2020}} &\textbf{Visual:} VGG19 &\multirow{2}{100pt}{MLP}    &Getty Image & Accuracy: 93.6\%  \\
    &\textbf{Textual:} GloVe  &  &VSO\_VT & Accuracy: 87.2\%  \\
    \hline
    \multirow{2}{70pt}{Jiang et al. \cite{Jiang2020}} &\textbf{Visual:} ResNet152 &\multirow{2}{60pt}{Soft-max} & MVSA-single &Accuracy: 74.21\% \\
    &\textbf{Textual:} GloVe and Bert &  &MVSA-multiple &Accuracy: 71.46\% \\
    \hline
    \multirow{3}{*}{Zhao et al. \cite{Zhao2019}} & Visual: Low-, Mid- and High-level Feature & \multirow{3}{*}{SVM} &\multirow{3}{*}{VSO} &\multirow{3}{*}{Accuracy: 87\%} \\
    &\textbf{Textual:} Word2vec &  &  &  \\
    &\textbf{Social:} Lifespan, Emotion and Reach &  &  &  \\
    \hline
    \multirow{2}{70pt}{Guo et al. \cite{Guo2021}} &\textbf{Visual:} ResNet152 Pre-trained on Place365 and ImageNet &\multirow{2}{60pt}{Fully connected}  &RON &Avg. Accuracy: 53.51\% \\
    &Textual: GloVe &  & DMON &Avg. Accuracy: 80.81\% \\
    \hline
    \multirow{4}{70pt}{Huang et al. \cite{Huang2020}} &\textbf{Visual:} VGG-19. &\multirow{4}{60pt}{Soft-max}    &Twitter-W & Accuracy: 79\% \\
    &\multirow{3}{*}{\textbf{Textual:} GloVe}    &  &Getty Images -W & Accuracy: 88\% \\
    &  &  & Flickr-W & Accuracy: 87\% \\
    &  &  & Flickr-M & Accuracy: 89\% \\
    \hline
    \multirow{3}{70pt}{You et al. \cite{You2016}} &\textbf{Visual:} Pre-trained CNN &\multirow{3}{60pt}{Soft-max} &Getty Images &Accuracy: 80\% \\
    & \multirow{2}{100pt}{\textbf{Textual:} Word2vec was initially employed and then taking the average} &  &Twitter & Accuracy: 80.9\% \\
    &   &   &AMT-Twitter & Accuracy: 76.9\% \\
    &   &   &   \\
    &   &   &   \\
    \hline
    \multirow{2}{70pt}{Duong et al. \cite{Duong2017}} &\textbf{Visual:} Inception &\multirow{2}{*}{NNs}  &Flicker &Accuracy: 93.44\%   \\
    &\textbf{Textual:} GloVe  &  & Reddit &Accuracy: 86.92\% \\
    \hline
    \multirow{3}{70pt}{Xu et al. \cite{Xu2019a}} &\textbf{Visual:} VGG19 & \multirow{3}{*}{MLP} & Flicker & Accuracy: 85.9\% \\
    &\textbf{Textual:} GloVe &  &Twitter &Accuracy: 76.7\% \\
    &\textbf{Network:} network embedding method &  & Flicker-Ml & Accuracy: 88.1\% \\
    \hline
    \multirow{2}{70pt}{Xu et al. \cite{Xu2021}} &\textbf{Visual:} VGG19 &\multirow{2}{*}{MLP} &Flickr & Accuracy: 87.1\% \\
    &\textbf{Textual:} GloVe &  &Getty Image &Accuracy: 87.8\% \\
    \hline
    \multirow{2}{70pt}{Zhu et al. \cite{Zhu2022}} &\textbf{Visual:} Faster R-CNN and ResNet-101 &\multirow{2}{60pt}{Soft-max} &MVSA-single &Accuracy: 75.19\% \\
    &\textbf{Textual:} BERT-Base &  & MVSA-multiple & Accuracy: 73.52\% \\
    \hline
    \multirow{2}{70pt}{Tashu et al. \cite{Tashu2021}} &\textbf{Visual:} ResNet50 &\multirow{2}{60pt}{Soft-max Classifier}   &\multirow{2}{60pt}{Wiki-Art Emotions} &\multirow{2}{*}{Accuracy: 77.3\%} \\
    &\textbf{Textual:} GloVe and Emotion Category extracted using three layer FFNN &    &  &  \\
    \hline
    \multirow{2}{*}{Ortis et al. \cite{Ortis2021}} &\textbf{Visual:} RGB histogram, GIST, BOW, GoogLeNet, DeepSentiBank, and Places205 &\multirow{2}{*}{Linear SVM} &\multirow{2}{*}{Flickr} & \multirow{2}{95pt}{73.96 ±0.39\% \&72.66 ±0.70\% represents the average and STD} \\
    &\textbf{Textual:} BOW, SentiwordNet, and word dictionary &  &  &  \\
    \hline
    \multirow{2}{70pt}{Cai and Xia \cite{Cai2015}} &\textbf{Visual:} CNN &\multirow{2}{60pt}{Soft-max} &Twitter (TD1) & Accuracy: 78\% \\
    &\textbf{Textual:} Word2vec & &Twitter (TD2) &Accuracy: 79.6\% \\
    \hline
    \multirow{4}{70pt}{You et al. \cite{You2016a}} &\textbf{Visual:} VGG19. &\multirow{4}{60pt}{Soft-max}  &Getty & Accuracy: 90.2\% \\
    & \multirow{3}{100pt}{\textbf{Textual:} One-Hot Encoding, GloVe} &  &Twitter &Accuracy: 96.4\% \\
    &  &  &Twitter AMT & Accuracy: 90.4\% \\
    &  &  &VSO-VT & Accuracy: 83.3\% \\
    \hline
    \multirow{3}{70pt}{Yang et al. \cite{Yang2021a}} &\textbf{Visual:} VGG-Object and VGG-Place &\multirow{3}{60pt}{MLP and a stacking-pooling module} &MVSA-single &Accuracy: 72.98\% \\
    &\multirow{2}{*}{\textbf{Textual:} GloVe} &  & MVSA-multiple &Accuracy: 72.36\% \\
    &  &  &Tum-Emo &Accuracy: 66.46\% \\
    \hline
    \multirow{4}{70pt}{Yadav and Vishwakarma \cite{Yadav2022}} &\textbf{Visual:} Inception V3 &\multirow{4}{60pt}{Soft-max}  &MVSA-single &Accuracy: 79.47\% \\
    & \multirow{3}{*}{\textbf{Textual:} GloVe} &  &MVSA-multiple &Accuracy: 77.89\% \\
    &  &  &Flickr &Accuracy: 89.30\% \\
    &  &  &Getty Image &Accuracy: 92.65\% \\
    \hline
    \multirow{2}{70pt}{Miao et al. \cite{Miao2018}} &\textbf{Visual:} VGG19 & \multirow{2}{*}{Dense Layer}  &\multirow{2}{*}{MM\_IMDB} &\multirow{2}{*}{Accuracy: 64.2\%} \\
    &\textbf{Textual:} GloVe  &  &  &  \\
    \hline
    \multirow{2}{70pt}{Zhu et al. \cite{Zhu2019}} &\textbf{Visual:} Inception-V3 &\multirow{2}{60pt}{Two-layer Perceptron}  &Getty Images &Accuracy: 91.3\% \\
    &\textbf{Textual:} GloVe &  & VSO & Accuracy: 85.1\% \\
    \hline
    \multirow{2}{70pt}{Xu and Mao \cite{Xu2017}} &\textbf{Visual:} VGG19-ImageNet \& VGG19-Place &\multirow{2}{60pt}{Soft-max} &MVSA-single &Accuracy: 69.84\% \\
    &\textbf{Textual:} GloVe  &  & MVSA-multiple & Accuracy: 68.86\% \\
    \hline
    \multirow{3}{70pt}{Yang et al. \cite{Yang2021}} &\textbf{Visual:} YOLOv3, VGG-Place, and ResNet used for object and scene memory banks &\multirow{3}{60pt}{Soft-max Classifier} & MVSA-single & Accuracy: 73.77\% \\
    & \multirow{2}{*}{\textbf{Textual:} GloVe} &  &MVSA-multiple & Accuracy: 72.49\% \\
    &  &  &Tum-Emo & Accuracy: 66.72\% \\
    \hline
    \multirow{5}{70pt}{Cao et al. \cite{Cao2020}} &\textbf{Visual:} local (Region Proposal Network) and Global (ResNet101) &\multirow{5}{60pt}{Soft-max}  &Twitter-W & Accuracy: 82.4\% \\
    &\multirow{5}{100pt}{\textbf{Textual:} GloVe and LSTM for sentence representation}   & &Getty Images-W. &Accuracy: 85.6\% \\
    &  &  & Flickr-W. & Accuracy: 84.1\% \\
    &  &  & VSO- Strong. & Accuracy: 90.5\% \\
    &  &  & &   \\
    \hline
    \multirow{3}{70pt}{Ye et al. \cite{Ye2019}} &\textbf{Visual:} ResNet101 &\multirow{3}{60pt}{Soft-max}  & PR-150K & Accuracy: 73.3\% \\
    &\multirow{2}{100pt}{\textbf{Textual:} Neural Network Embedding layer} &  &MVSO  &Accuracy: 80.1\% \\
    &  &  & VSO & Accuracy: 85.6\% \\
    &   &   &   &   \\
    \hline
    \multirow{4}{70pt}{Xu et al. \cite{Xu2019}} &\textbf{Visual:} Local (Faster R-CNN) and Global (VGG19) &\multirow{4}{60pt}{Soft-max}  & Flickr-W &Accuracy: 84.9\% \\
    & \multirow{3}{*}{\textbf{Textual:} GloVe} &  &Flickr-ML &Accuracy: 87.8\% \\
    &  &  &Getty Images-W & Accuracy: 86.5\% \\
    &  &  &Flickr-IML & Accuracy: 83.1\% \\
    \hline
    \multirow{2}{70pt}{Liao et al. \cite{Liao2022}} &\textbf{Visual:} EfficientNet-b0 &\multirow{2}{60pt}{Soft-max}  &MVSA-single &Max accuracy: 73.84\% \\
    &\textbf{Textual:} GloVe and text level Graph Neural Network (GNN) &  &Twitter26k & Max accuracy: 93.88\% \\
    \hline
    \multirow{2}{70pt}{Truong and Lauw \cite{Truong2019}} &\textbf{Visual:} VGG16 &\multirow{2}{60pt}{soft-max} &\multirow{2}{70pt}{MM Yelp} &\multirow{2}{95pt}{Accuracy: 61.88\%} \\
    &\textbf{Textual:} GloVe. &  &  &  \\
    \hline
\end{longtable}

On the other hand, late fusion is used to integrate the outputs of several different sentiment classifiers trained independently. Kumar et al. \cite{Kumar2020} proposed a hybrid DL model for fine-grained sentiment prediction in multimodal data using DL networks and ML to handle two distinct systems (i.e. textual and visual) and their combination inside online material utilising decision-level fusion. The proposed contextual ConvNet–SVMBoVW model was trained using comments and postings (text, picture and infographic) generated using the \#CWC2019 hashtag on Instagram and Twitter. The proposed model's accuracy is superior to the text and picture modules. Zhang et al. \cite{Zhang2022} proposed a hybrid fusion network to collect inter- and intra-modal characteristics. A multi-head visual attention was presented to derive correct meaning and sentimental insights from text contents, guided by visual cues. In the decision fusion step, several baseline classifiers were trained to learn distinct discriminative knowledge from numerous modal representations. The final decision was made by combining decision supports from the baseline classifiers. Huang et al. \cite{Huang2019} presented a new visual-textual SA model called deep multimodal attentive fusion, which uses visual and semantic internal correlations to analyse sentiments. Two unimodal attention models were presented to acquire effective emotion classifiers for text and image modality. Then, a multimodal attention model was developed to implement the intermediate fusion for joint sentiment classification. Finally, late fusion was used to merge the three attention models. Kumar and Garg \cite{Kumar2019} suggested a multimodal SA approach to analyse visual-textual data (or infographic and typographic). A CNN was used to score the image sentiment utilising SentiBank and SentiStrength (R-CNN), and an innovative hybrid (lexicon and ML) method was used to score text sentiment. After separating text from images using optical character recognition, multimodal sentiment scoring was performed by combining the sentiment ratings from images and texts. High accuracy was achieved in the suggested model's performance using a random multimodal tweet dataset. Yu et al. \cite{Yu2016} suggested a CNN-based visual–textual SA system. They built CNN models for Chinese microblogging that combined picture and text characteristics using written and visual information. Zhang et al. \cite{Zhang2018} proposed a quantum-inspired multimodal SA (QMSA) system. The system contains a quantum-inspired multimodal representation model. This approach addressed the semantic gap using a multimodal decision fusion strategy motivated by quantum interference and a density matrix to define the interactions across various modalities. Liu et al. \cite{Liu2020} suggested a useful method for the SA of GIF videos with written descriptions that combine visual and linguistic information. Each GIF video's visual characteristics were retrieved and categorised using a sentiment classifier. The sentiment probability obtained from the visual classifier was further transformed into a sentiment score by developing a mapping function. Simultaneously, the SentiWordNet 3.0 model was used to assign a sentiment score to the extracted sets of significant sentiment words from the attached textual annotations. Finally, a sentiment score function comprising visual and textual components was built with a remarkable difference threshold to improve the fused sentiment score further. Qian et al. \cite{Qian} used deep CNN trained on Twitter data to provide a unique approach for extracting features from Twitter photos and the accompanying labels or tweets. They fine-tuned AlexNet to extract the visual features and the affective space of English ideas, which were employed to extract the text features. Finally, a unique sentiment score was suggested to integrate the picture and text predictions. The assessment was based on a Twitter dataset, including photos, labels and tweet messages. Kumar et al. \cite{Kumar2021} proposed a visual–textual emotion classification model using DL based on fusion strategies. Firstly, they performed a joint fusion to combine the visual and textual data. Secondly, a decision fusion model was created to combine the results of the textual, visual and joint fusion outputs. A new emotional dataset was generated using the B-T4SA dataset. Table \ref{tbl12} summarises the literature for visual–textual SA using late fusion.

\begin{longtable}{p{60pt}p{95pt}p{90pt}p{80pt}p{80pt}}
    \caption{Summary of the literature using late fusion} \label{tbl12}\\
    \hline
    \textbf{Author}  &\textbf{Features}   &\textbf{Classification Method} &\textbf{Dataset}    &\textbf{Performance Metric} \\
        \hline
        \multirow{4}{60pt}{Huang et al. \cite{Huang2019}} &\textbf{Visual:} VGG19 &\textbf{Visual model:} Fully connected layer &Twitter-W &Accuracy: 76.3\% \\
        &\textbf{Textual:} GloVe &\textbf{Textual model:} Fully connected layer   &Getty Images-W &Accuracy: 86.9\% \\
        &  &\textbf{Joint Model:} Fully connected layer  &Flickr-W &Accuracy: 85.9\% \\
        &  &\textbf{Late:} Custom defined rules   &Flickr-M &Accuracy: 88\% \\
        \hline
        \multirow{3}{60pt}{Kumar et al. \cite{Kumar2020}} &\textbf{Visual:} BOVW using LBP Descriptor  &\textbf{Visual model:} SVM  &\multirow{3}{80pt}{Tweets collected using \#CWC2019 on Instagram and Twitter} &\multirow{3}{*}{Accuracy: 91\%} \\
        &\textbf{Textual:} GloVe, Vadar &\textbf{Textual model:} Aggregation score between (CNN and Senticircle). &  &  \\
        &   &\textbf{Late:} Boolean operations &    &   \\
        \hline
        \multirow{3}{60pt}{Kumar and Garg \cite{Kumar2019}} &\textbf{Visual:} R-CNN, Sentibank  &\textbf{Visual model:} hybrid of SentiBank and RCNN    &\multirow{3}{80pt}{Collected tweets for LGBT verdict of Indian Penal Court (IPC) section 377.} &\multirow{3}{*}{Accuracy: 91.32\%} \\
        &\textbf{Textual:} BOW, Unigrams, PoS, Negation, Count of Emoticon, Count of elongated words, Count of capitalised words, Length of message &\textbf{Textual model:} Aggregation score between (Gradient Boosting, SentiCircle) &  &  \\
        &   &\textbf{Late:} Aggregating the Sentiment Polarity and scores   &   &   \\
        \hline
        \multirow{3}{60pt}{Yu et al. \cite{Yu2016}} &\textbf{Visual:} CNN &\textbf{Textual model:} LR   &\multirow{3}{80pt}{Sina Weibo} &Accuracy / 2 class: 81.1\% \\
        &\textbf{Textual:} Word2vec &\textbf{Visual model:} LR  &  &Accuracy / 3 class: 74.8\% \\
        &  &\textbf{Late:} Average strategy   &   &   \\
        \hline
        \multirow{4}{60pt}{Zhang et al. \cite{Zhang2018}} &\textbf{Visual:} BOVW using SIFT Descriptor &\multirow{4}{85pt}{RF, SVM, Quantum-inspired multimodal SA for late fusion}   &\multirow{2}{80pt}{Getty Images} &Accuracy / RF: 88.24\% \\
        & &  &  &Accuracy / SVM: 79.76\% \\
        &\textbf{Textual:} GloVe  &  &\multirow{2}{80pt}{Flicker}  &Accuracy / RF: 93.14\% \\
        &  &  &  & Accuracy / SVM: 92.43\% \\
        \hline
        \multirow{3}{60pt}{Liu et al. \cite{Liu2020}} &\textbf{Visual:} VGG16 for the sequence level feature with C3D for the frame level features &\textbf{Visual model:} CONVLSTM then Softmax &T-GIF Dataset: &Accuracy: 78.39\% \\
        &\multirow{2}{90pt}{\textbf{Textual:} Synset Forest} &\textbf{Textual model:} SentiWordNet3.0 model  &GSO-2016 Dataset &Accuracy: 75.13\% \\
        &  &\textbf{Late:} Weighted sum for late fusion  &Adjusted-GIFGIF Dataset &Accuracy: 74.03\% \\
        \hline
        \multirow{3}{60pt}{Qian et al. \cite{Qian}} & \textbf{Visual:} AlexNet &\textbf{Visual model:} Softmax  &\multirow{2}{*}{Twitter dataset} &\multirow{3}{*}{Accuracy: 80.51\%} \\
        &\textbf{Textual:} AffectiveSpace & \textbf{Textual model:} SVM &  &  \\
        &   &\textbf{Late:} Weighted sum &   &   \\
        \hline
        \multirow{5}{60pt}{Zhang et al. \cite{Zhang2022}} &\textbf{Visual:} VGG16 &\textbf{Visual model:} Fully connected layer  &MM Yelp   &Avg. Accuracy: 63.41\% \\
        &\textbf{Textual:} Bert &\textbf{Textual model:} Fully connected layer  &CMU-MOSII  &Avg. Accuracy: 35.42\%  \\
        &   &\textbf{Joint Model:} Fully connected layer   &CMU-MOSEI   &Avg. Accuracy: 51.65\% \\
        &  &\textbf{Late:} Classification based approach using NN   &Twitter-15 &Avg. Accuracy: 78.62\%   \\
        &   &   &Twitter-17   &Avg. Accuracy: 71.35\% \\
        \hline
        \multirow{4}{60pt}{Kumar et al. \cite{Kumar2021}}   &\textbf{Visual:} VGG16. &\textbf{Visual model:} Soft-max  &Emotion  &Accuracy: 90.20\% \\
        &\textbf{Textual:} GloVe    &\textbf{Textual model:} Soft-max    &B-T4SA &Accuracy: 86.70\%    \\
        &   &\textbf{Joint Model:}  Soft-max &   &   \\
        &   &\textbf{Late:} Weighted average  &   &   \\
    \hline
\end{longtable}

\subsubsection{VISUAL–TEXTUAL DATASETS} \label{subsubsec:visual_textual_datasets}
With a large number of social media networks (e.g. Twitter, Flicker, Facebook and Instagram) where people can share their opinions and sentiments on various topics, a large dataset for visual–textual contents may be collected. However, collecting such a large number of datasets requires significant effort for annotation. Some benchmark datasets available for visual–textual SA are presented in Table \ref{tbl10:most_popular_dataset_visual-textual}.

\begin{table}
    \caption{Most popular datasets used in visual–textual SA}
    \label{tbl10:most_popular_dataset_visual-textual}
    \begin{tabular}{p{80pt}p{370pt}}
    \hline
    \textbf{Dataset}    &\textbf{Description  and Link}   \\
    \hline
    Photo Tweet Sentiment Benchmark &A small dataset having 603 tweets containing  both images and their related text descriptions used to cover a wide range of topics. It was initially provided by Borth et al. \cite{Borth2013} which was then labelled by Amazon Mechanical Turk (AMT) workers, yielding 470 positive and 133 negative categories. \url{https://www.ee.columbia.edu/ln/dvmm/vso/download/twitter_dataset.html}   \\
    \hline
    VSO: Image Dataset  &The database consists of two datasets: a collection of Flickr photos with Creative Commons licenses that were used to train/test 1,200 ANP detectors in SentiBank and a set of images connected with the entire VSO, which includes 3,244 ANPs \cite{Borth2013}. \url{https://www.dropbox.com/sh/xs049whsb3wt3c8/AAB5TSbydN-6yGqYnHo1XHgCa?dl=0}    \\
    \hline
    Multi-view SA Dataset (MVSA) &An MVSA for visual–textual SA was recently provided by Niu et al. \cite{Niu2016}. Only tweets with text, as well as an image that had observable photos, were preserved for annotation. There are three sentiments in 4,869 texts (positive, negative and neutral). \url{https://mcrlab.net/research/mvsa-sentiment-analysis-on-multi-view-social-data/}  \\
    \hline
    T4SA    &A huge dataset of over 3 million tweets provided by Vadicamo et al. \cite{Vadicamo2017} contains text and images categorised  according to the polarity of the text's sentiment (negative = 0, neutral = 1 and positive = 2). \url{http://www.t4sa.it/} \\
    \hline
    \end{tabular}
\end{table}
\subsubsection{EVALUATION MEASURES} \label{subsubsec:evaluation_measures}
Various measures are utilised to analyse and compare the performance and the efficiency of the model. Confusion matrices (also known as error matrices or truth tables) are a well-known method for assessing and analysing  the effectiveness of a categorisation model. This matrix displays the number of properly and wrongly categorised samples identified by a classifier. The abbreviations TP (True Positive), FP (False Positive), FN (False Negative), and TN (True Negative) in the confusion matrix correspond to the following:
\begin{itemize}
    \item TP: Number of instances where the expected class label is positive while the actual class label is correct.
    \item FP: Number of instances where the expected class label is positive and the actual class label is incorrect.
    \item FN: Number of instances where the expected class label is negative and the actual class label is incorrect.
    \item TN: Number of samples where the expected class label is negative and the actual class label is correct.
\end{itemize}

Accuracy, Recall, Precision and the F1-score are regularly used performance assessment metrics by many researchers, which are determined using the confusion matrix. Accuracy is the ratio of accurately predicted instances to the total number of examples. Its value is determined using Eq. \ref{eq:accuracy}.

\begin{equation}
    Accuracy = \frac{TP + TN}{Tp + TN + FP + FN}
    \label{eq:accuracy}
\end{equation}

Precision refers to the total number of expected class labels that are correct for each class. The precision value is determined using Eq. \ref{eq:precision}.

\begin{equation}
    Precision = \frac{TP}{TP + FP}
    \label{eq:precision}
\end{equation}

Recall value is the weighted average of labels that have been correctly categorized for each class. This value is determined using Eq. \ref{eq:recall}.

\begin{equation}
    Recall = \frac{TP}{TP + TN}
    \label{eq:recall}
\end{equation}

F1-score is utilized to integrate precision and recall scores into one measurement. This number ranges from 0 to 1, and if the classifier properly identifies every sample, it delivers a value of 1, indicating a high level of classification success. The F1-score value is determined using Eq. \ref{eq:f1}.

\begin{equation}
    F1-score = \frac{2 \times precision \times recall}{precision + recall}
    \label{eq:f1}
\end{equation}

Accuracy is often the most used statistic for measuring algorithm performance. However, other measures (accuracy, recall, and F1-score) were utilized by researchers to evaluate their investigations. For instance, Xu et al. \cite{Xu2019} and Liu et al. \cite{Liu2020} evaluated their models using accuracy, precision, recall, and F1-score. In contrast, Kumar and Garg \cite{Kumar2019} evaluated the efficacy of their model using an accuracy metric, while Zhu et al.  \cite{Zhu2022} employed both accuracy and F1-score for performance assessment.
\section{CHALLENGES} \label{sec:challenges}
Multimodal SA is the process of extracting emotions and sentiments associated with multimodal inputs. Nowadays, the majority of social media posts and blogs are multimodal (i.e. more than one modality is linked with the data), consisting primarily of textual and visual elements. Although plenty of advancements have been achieved in this area, there are still a lot of problems and challenges that need to be addressed to enhance the model's performance. Some of the challenges in multimodal SA are as follows.
\begin{itemize}
    \item Heterogeneity in Feature Space: Different from traditional single-modality SA, multimodal SA includes a varied collection of manifestation patterns. Visual material and textual description are diverse in feature spaces. As a consequence, SA should be effective in bridging the gap between diverse modalities \cite{Yang2021a,Yadav2022,Huang2019}.
    \item Extraction of the Sentiment Intensity: Different semantic information may be covered by the visual content and word description that comes in various forms, such as expressions, emoticons and gestures. The present computational approaches are primarily concerned with extracting syntactic information from user content; however, they fail to capture the sentiment information, which is critical in the SA process \cite{Zhang2018}.
    \item High-dimensional Feature Space: As multiple modalities are involved, each modality has a distinguishing property that aids in characterising it. As a result, the same strategies for feature extraction cannot be used to all modalities (i.e. textual and visual). To categorise sentiments in visual–textual SA, feature vectors for each modality are created individually, and it is then combined into one global vector, which is a time-consuming procedure, leading to a high-dimensional feature vector that must be solved \cite{Zhao2019,Xu2019a}. 
    \item Cross-modal Relationships: For SA, a huge amount of data is accessible through the social platforms using different formats. Users express their opinions/sentiments in several modes concurrently because they aim to transmit the same polarity of feeling across various modes. As a result, the relationship between visual content and its textual semantics should be considered to achieve better sentiment classification \cite{Zhang2021,You2016,Peng2022}. 
    \item Incomplete Modality: It is not uncommon for one modality to be absent from multimodal data. Many users will often submit tweets without including photos, whereas photographers will frequently share photos without including a written description. Handling incomplete multimodal data for sentiment classification is another problem \cite{Duong2017}.
    \item Domain-independent Sentiment Polarity: Most present research attempts to predict sentiment polarity using domain knowledge. The system cannot infer sentiments from microblogs if trained on plain text content from documents. This characteristic prevents it from adapting to new domains \cite{Dragoni2018}.
    \item Integrating Data from Several Sources: The fusion approach should be used to integrate the information gathered from all modalities. It is crucial to carefully consider which modality contributes most to the fusion by allocating weights to each modality throughout the fusion process. In feature-level fusion, the feature vectors with the largest weighted modality will be the dominant element in the fusion process. Moreover, the sentiment score of the modality with the greatest weight should be included in decision-level fusion \cite{Huang2020,Zhang2021,You2016}.
    \item Dataset Scarcity: There exists a shortage of available datasets, especially in multimodal SA using visual and textual content, and most of the existing datasets suffer from many problems. One of these problems is an imbalance in the class distribution, which will impact how well ML-based sentiment prediction performs. Another issue is the subjectivity of human emotion, which causes label unreliability. Even manually annotated sentiment labels are not 100\% trustworthy \cite{Borth2013,Truong2019}.
    \item Noise Presented with Additional Modalities: Images and words may be quite semantically expressive. In most circumstances, the usable semantic segment for SA is only a portion of the picture or text. The remainder is redundant and may lead to sentiment misinterpretation errors. The rising volume of data in visual–textual SA exacerbates this difficulty. Misunderstanding one modality might lead to poor results even when the other is correctly assessed. A unique method is required to combat multimodal noise \cite{Zhou2020}.
\end{itemize}
\section{APPLICATIONS OF SENTIMENT ANALYSIS} \label{sec:applications}
The increasing availability of emotional data from numerous forums, blogs and social networks has raised academic and industry awareness of SA. SA may help organisations understand people’s attitudes and preferences based on prior behaviour. Thus, it may enable them to personalise their goods and services to their needs. The various application fields of SA are as follows.
\begin{itemize}
    \item Predictions for the Box Office. With the fast growth of social media, many online film reviews are accessible in text and video format, which enables forecasting the box office success of films. A simple sentiment-aware autoregressive model was presented by Nagamma et al. \cite{Nagamma2015}, which used TF-IDF values as features and Fuzzy Clustering. An SVM classifier was also developed for forecasting box office revenue trends based on review sentiment. Timani et al. \cite{Timani2019}, created three sentiment indexes using YouTube movie trailer comments to quantify the sentiment of movie reviews. These indexes were then used to predict a movie’s box office collection. The comments on the trailers assisted distributors and filmmakers in estimating the movie’s reaction rate.
    \item Forecasting the Stock Market. Despite the computer world's advancement, the stock market's unpredictable characteristic makes forecasting one of the most challenging undertakings. SA may assist in the development of efficient methods. In Moghaddam et al. \cite{Moghaddam2016}, ANNs, which were trained via backpropagation, were used to forecast the NASDAQ stock market index.
    \item Business Intelligence. Many firms are now using SA to aid in decision-making and business improvement. In Rokade and Aruna Kumari \cite{Rokade2019}, a revolutionary approach was proposed to business analytics in modern enterprises.
    \item Summarisation of Television Programmes and Newscasts. A system that uses multimodal SA to automatically assess broadcast video news and construct summaries of television shows was proposed by Ellis et al. \cite{Ellis2014}. They described a method for analysing and creating person-specific fragments from news videos, delivering 929 sentence-length videos annotated using Amazon Mechanical Turk.
    \item Recommender Systems. Numerous applications provide suggestions based on users’ prior experience. For instance, in retail, consumers seeking a certain product may obtain suggestions for future endeavours. Zheng et al. \cite{Zheng2015} presented a hybrid technique that offers correct recommendations.
    \item Prediction of Political Trends. The growing popularity of social media platforms improves the potential of forecasting the result of an election. For example, Chauhan et al. \cite{Chauhan2021} wrote a survey article that discusses the assessment of SA methodologies and attempts to demonstrate the researchers’ contribution to forecasting election outcomes using social media information. Moreover, this article recommends future research on election prediction using social media information.
    \item Healthcare. It is the most frequently utilised domain, used to assess patient evaluations regarding their health that are shared on different social media sites.  SA model was developed by Ramirez-Tinoco et al. \cite{Ramirez-Tinoco2019} to obtain sentiment and emotional information that may aid healthcare practitioners in comprehending their patients’ emotions and concerns by taking appropriate remedial action.
\end{itemize}
\section{CONCLUSION} \label{sec:conclusion}
This paper provides an overview of visual–textual SA and their related techniques by reviewing the current works to provide researchers with a complete understanding of the methodology and resources available for visual and textual SA. It also covers categorising and summarising the most widely used SA methodologies and their benefits and limitations. It includes important procedures, such as pre-processing, feature extraction, data fusion strategies and the most widely used sentimental datasets. It also examines some of the field’s most pressing challenges and applications.

Our investigation of this field reveals its potential for leveraging complementary information channels in SA, often surpassing the performance of unimodal approaches. Additionally, it can enrich other methods, such as entity recognition and subjectivity analysis, which can directly benefit from unimodal SA. Our review paper will promote further interdisciplinary cooperation in this developing sector. Future research may include a more in-depth examination of sentiment assessment in diverse domains. Numerous case studies may be undertaken to evaluate the efficacy of different methodologies in exploring sentiments.

\section*{Declarations}

\begin{itemize}
\item \textit{Funding:} No funds, grants, or other support was received.
\item \textit{Conflict of interest:} The authors have no conflicts of interest to declare that are relevant to the content of this article.
\item \textit{Availability of data and materials:} Data sharing is not applicable to this article as no new data were created or analyzed in this study.
\end{itemize}


\bibliographystyle{unsrt}  
\bibliography{references}

\end{document}